\documentclass[10pt,twocolumn,letterpaper]{article}

\usepackage[algorithms]{wacv}      %

\usepackage{graphicx}
\usepackage{amsmath}
\usepackage{amssymb}
\usepackage{booktabs}
\usepackage{amsthm}

\usepackage{enumitem}
\usepackage[ruled]{algorithm2e}
\usepackage{multirow}
\usepackage{float}
\usepackage{booktabs}
\usepackage{xcolor}
\usepackage{blindtext}
\setitemize{noitemsep,topsep=0pt,parsep=0pt,partopsep=0pt}
\newcommand{\scorenp}{F1-Score\textsubscript{non-private}}
\newcommand{\scoredp}{F1-Score\textsubscript{DP}}
\newcommand{\scorefl}{F1-Score\textsubscript{FedAVG}}
\newcommand{\scorefldp}{F1-Score\textsubscript{FeAm-DP}}
\newcommand{\robertabase}{RoBERTa\textsubscript{base}}

\newcommand{\layoutlm}{LayoutLMv3~\cite{kie-layoutlmv3-Huang2022}}
\newcommand{\layoutlmbase}{LayoutLMv3\textsubscript{base}}
\newcommand{\layoutlmbasetsft}[1]{\layoutlmbase\textsubscript{-TSP-#1}}
\newcommand{\layoutlmbasec}{LayoutLMv3\textsubscript{base-chinese}}

\newcommand{\downarrowc}{\textcolor{red}{\downarrow}}
\newcommand{\uparrowc}{\textcolor{green}{\uparrow}}
\newtheorem{definition}{Definition}
\newtheorem{guidelines}{Guideline}
\usepackage[pagebackref,breaklinks,colorlinks]{hyperref}

\usepackage[capitalize]{cleveref}
\crefname{section}{Sec.}{Secs.}
\Crefname{section}{Section}{Sections}
\Crefname{table*}{table*}{Tables}
\crefname{table*}{Tab.}{Tabs.}
\usepackage{xcolor}
\usepackage{soul}
\newcommand{\mathcolorbox}[2]{\colorbox{#1}{$\displaystyle #2$}}
\SetKwComment{Comment}{/* }{ */}
\SetKwInput{KwInit}{Init}
\SetKwFor{ForEachParallel}{for}{in parallel do}{end}
\SetKwFor{ForEachGradient}{for each}{compute}{end}
\SetKwBlock{Server}{Server Executes:}{end}
\SetKwBlock{Client}{ClientUpdate}{end}

\def\wacvPaperID{652} %
\def\confName{WACV}
\def\confYear{2024}

\begin{document}
	
	\title{PrIeD-KIE: Towards Privacy Preserved Document Key Information Extraction}
	
	\author{Saifullah Saifullah$^{12}$, Stefan Agne$^{23}$, Andreas Dengel$^{12}$, Sheraz Ahmed$^{23}$ \\
		$^1$Department of Computer Science, University of Kaiserslautern-Landau\\
		Erwin-Schrödinger-Straße 52, 67663, Kaiserslautern, Rhineland-Palatinate, Germany\\
		$^2$German Research Center for Artificial Intelligence GmbH\\
		Trippstadter Straße 122, 67663, Kaiserslautern, Rhineland-Palatinate, Germany\\
		$^3$DeepReader GmbH, 67663 Kaiserlautern, Germany\\
		{\tt\small firstname.lastname@dfki.de}
}
\maketitle

\begin{abstract}
	In this paper, we introduce strategies for developing private Key Information Extraction (KIE) systems by leveraging large pretrained document foundation models in conjunction with differential privacy (DP), federated learning (FL), and Differentially Private Federated Learning (DP-FL). 
	Through extensive experimentation on six benchmark datasets (FUNSD, CORD, SROIE, WildReceipts, XFUND, and DOCILE), we demonstrate that large document foundation models can be effectively fine-tuned for the KIE task under private settings to achieve adequate performance while maintaining strong privacy guarantees.
	Moreover, by thoroughly analyzing the impact of various training and model parameters on model performance, we propose simple yet effective guidelines for achieving an optimal privacy-utility trade-off for the KIE task under global DP. 
	Finally, we introduce FeAm-DP, a novel DP-FL algorithm that enables efficiently upscaling global DP from a standalone context to a multi-client federated environment. We conduct a comprehensive evaluation of the algorithm across various client and privacy settings, and demonstrate its capability to achieve comparable performance and privacy guarantees to standalone DP, even when accommodating an increasing number of participating clients.
	Overall, our study offers valuable insights into the development of private KIE systems, and highlights the potential of document foundation models for privacy-preserved Document AI applications.
	To the best of authors' knowledge, this is the first work that explores privacy preserved document KIE using document foundation models.
\end{abstract}

\section{Introduction}
\label{sec:intro}

In today's digital landscape, the advent of AI-powered document processing systems has revolutionized the way businesses handle personal data~\cite{bert-devlin2019,kie-layoutlmv3-Huang2022,kie-tilt-Powalski2021,kie-formnet-Lee2022,doc-cls-Ferrando2020,layout-shen2021}, yet it has also raised concerns about privacy and data security~\cite{dp-Li2021,dp-Hoory2021,local-mdp-Feyisetan2019}.
Business-critical documents, such as invoices, contracts, financial reports, and customer databases, house a wealth of sensitive and confidential information, including names, addresses, contact details, social security numbers, and financial details.
As AI systems become increasingly integrated into document processing workflows, ensuring the privacy and security of this valuable data has become an imperative task. Increasing numbers of studies have demonstrated that machine learning (ML) models can be vulnerable to privacy threats~\cite{attacks-survey-Al-Rubaie2019,membership-inf-att-Shokri2017}. Inadvertent memorization~\cite{memorization-Carlini2019} of training data and the ability to reconstruct training samples~\cite{model-inversion-Coavoux2020,model-inversion-att-Fredrikson2015} within these models are just a few examples which pose substantial risks to user data privacy.
As a result, their adoption in industry and compliance with regulatory and ethical guidelines such as GDPR~\cite{GDPR2016a} and AI Act 2022 are significantly hindered.

To address the aforementioned issues, a diverse range of privacy-preserving approaches have recently emerged in the AI community such as Differential Privacy (DP)~\cite{dp-Dwork2006,dp-Dwork2014,dpsgd-Abadi2016}, Federated Learning (FL)~\cite{fl-McMahan2016,fl-rnn-McMahan2017LearningDP}, and data anonymization~\cite{priv-redaction-Sanchez2016}. In the context of Document AI, each of these approaches may have a different set of applications. As an example, DP may be applied to provide global privacy~\cite{dpsgd-Abadi2016,dp-Li2021,dp-lora-Yu2021}, where an organization provides machine learning-based services to external clients and intends to secure sensitive documents used during training, or local privacy~\cite{local-mdp-Feyisetan2019,local-dp-Lyu2020}, where individual clients are concerned about their personal information being leaked from documents uploaded to the service.
DP may also be combined with FL~\cite{fl-McMahan2016,fl-rnn-McMahan2017LearningDP}, where the model is trained across several participants, each optimizing the model locally on its own data, thus ensuring its privacy.
On the other hand, data anonymization through text redaction~\cite{priv-redaction-Sanchez2016} or sanitization~\cite{localdp-santext-YueDu21,local-dp-Xu2020} may also be used to directly introduce privacy into the document data.

In this paper, we address the privacy challenges in Key Information Extraction (KIE)~\cite{sroie-Huang2019,cord-park2019,funsd-jaume2019}, a vital task in modern document processing workflows that involves extracting targeted entities like names, addresses, and bank details from documents such as invoices, receipts, financial reports, and medical records. 
However, the sensitive nature of the data handled by KIE systems make them vulnerable to privacy attacks. 
Despite significant advances in multimodal KIE~\cite{kie-bertgrid-Denk2019,kie-vibertgrid,kie-layoutlmv2-xu2021,kie-layoutlmv3-Huang2022,kie-tilt-Powalski2021} and privacy preservation of textual documents~\cite{dp-Li2021,dp-Hoory2021,dp-Dupuy2022,local-mdp-Feyisetan2019,local-dp-Meehan2022}, research specifically tailored to privacy-enhancing approaches for unified KIE is lacking. To address this gap, this paper proposes a privacy-enabled KIE system by integrating unified document transformer models with privacy techniques. The contributions of this paper are three-fold:

\begin{itemize}
	\item[--] We propose strategies for building private KIE systems by utilizing large pretrained document foundation models in combination with differential privacy (DP), federated learning (FL), and Differentially Private Federated Learning (DP-FL) and demonstrate that given adequate hyperparameter tuning, multimodal transformer models can be fine-tuned on the KIE task with sufficient utility under strong privacy guarantees ($\epsilon\in\{8,20\}$ in case of DP and DP-FL).
	\item[--]  We perform extensive experiments for fine-tuning the models under global DP, and suggest simple but effective guidelines for achieving an optimal privacy-utility tradeoff.
	\item[--] We propose FeAm-DP, an algorithm for DP-FL training, which utilizes local client-level noise addition in combination with global Adam optimization to ensure privacy, and enables efficient upscaling of global DP from a standalone perspective to multiple clients.
\end{itemize}

\section{Related Work}
\subsection{Privacy in NLP and Document AI}
The fields of Natural language processing (NLP) and document AI are closely intertwined, and a variety of privacy approaches have emerged recently in both domains. 
Li \etal~\cite{dp-Li2021} investigated global DP for fine-tuning large pretrained language models and proposed practical guidelines tuning hyperparameters for text classification and generation tasks.
Hoory \etal~\cite{dp-Hoory2021} also studied global DP for the task of named-entity-recognition (NER) on medical documents and proposed a DP-based vocabulary for BERT \cite{bert-devlin2019} model. McMahan \etal~\cite{fl-rnn-McMahan2017LearningDP} demonstrated large-scale federated learning under client-level DP for textual data. Basu \etal~\cite{priv-documents-Basu2021} investigated DP in combination with federated learning for textual classification of financial documents. Several other studies have applied global DP~\cite{dp-Dupuy2022,dp-lora-Yu2021,dp-Wunderlich2022} as well as local DP~\cite{local-dp-Lyu2020,local-dp-Meehan2022,local-dp-Xu2020,localdp-santext-YueDu21,local-dp-cape-plant2021} to textual data. For a comprehensive review of these studies, we refer the reader to a related survey~\cite{priv-survey-Hu2023}.

Privacy for documents has been explored relatively little in the visual domain compared to the textual domain. In a comprehensive study, Saifullah \etal~\cite{privacy-doc-vis-Saifullah2022} investigated various privacy-preserving techniques such as DP, FL, and Secure Multi-Party Computation for the task of document image classification based on convolutional neural networks (CNNs). This work was also later extended in~\cite{saifullah2022privacy} to examine the effect of various privacy-preserving methods on the explainability of document image classifiers.

\subsection{Document Key Information Extraction}
Key Information Extraction (KIE) is a critical component of document analysis and has been extensively studied in the past~\cite{kie-bertgrid-Denk2019,kie-formnet-Lee2022,kie-layoutlmv3-Huang2022,kie-tilt-Powalski2021}. A variety of approaches have been explored in recent years for the KIE task, including grid-based methods~\cite{kie-chargrid-Katti2018,kie-bertgrid-Denk2019,kie-vibertgrid} that fuse textual information into the visual space, as well as transformer-based approaches~\cite{kie-layoutlmv2-xu2021,kie-layoutlmv3-Huang2022,kie-tilt-Powalski2021} that fuse textual, layout and visual information through self-attention~\cite{attention-vaswani2017} and use large-scale self-supervised pretraining for a unified document understanding. These techniques generally involve pretraining transformer models on large datasets and then fine-tuning them for downstream KIE tasks. The use of graphs to fuse multiple modalities has also been investigated recently for Document KIE~\cite{kie-formnet-Lee2022}.

\section{Preliminaries}
\subsection{Differential Privacy (DP)}
Differential Privacy (DP)~\cite{dp-Dwork2006,dp-Dwork2014} provides a formal definition of information release from private data. 
Applied to machine learning, DP provides strong privacy guarantees: severely limiting the possibility of using membership inference~\cite{membership-inf-att-Shokri2017}, model inversion~\cite{model-inversion-att-Fredrikson2015}, or linkage attacks~\cite{attacks-survey-Al-Rubaie2019} to learn about private training data through the publicly released model parameters. 
In this work, we mainly target the \textit{example-level privacy} for the KIE task under global approximate-DP (also known as $(\epsilon,\delta)$-DP) setting. Recall the formal definition of $(\epsilon,\delta)$-DP~\cite{dp-Dwork2006,dp-Dwork2014}:
\begin{definition}
	A randomized algorithm $\mathcal{M} : \mathcal{D} \rightarrow \mathcal{R}$ with domain $\mathcal{D}$ and range $\mathcal{R}$ is $(\epsilon,\delta)$-
	differentially private if for all adjacent datasets $D,D' \in \mathcal{D}$ and $R \subseteq \mathcal{R}$, $\mathbb{P}(\mathcal{M}(D)\in R)\leq e^{\epsilon}\mathbb{P}(\mathcal{M}(D')\in R)+\delta$
\end{definition}
Where the definition of adjacency of datasets depends on the task at hand. 
Similar to most prior works on \textit{example-level privacy} in machine learning~\cite{dpsgd-Abadi2016,dp-Li2021,dp-papernot2020making}, we use the following definition of adjacency: the two datasets $D,D'$ are considered adjacent if and only if $D'$ can be obtained from $D$ by adding or removing one record from $D$.
Intuitively speaking, applying DP to an algorithm $\mathcal{M}$ ensures that the outputs generated on similar inputs should be difficult to distinguish. Whereas, the degree of this indistinguishability is determined by the privacy parameters $\epsilon$ and $\delta$, with smaller values of $\epsilon$ and $\delta$ implying stronger privacy.

\subsubsection{Privatized Gradients via DP-SGD/Adam}
DP-SGD~\cite{dpsgd-Abadi2016} is the key algorithm for training a machine learning model for \textit{example-level privacy} under $(\epsilon,\delta)$-DP. The algorithm relies on adding noise to the per-example gradients of the model to ensure that the training process leaks limited information about the data it was trained on. In particular, for each training example, the per-example gradients are clipped to a constant bound $S$, and then gaussian noise $n\sim\mathcal{N}(0, \sigma^2S^2)$ is added to the sum of the gradients before the optimizer update. Where $\sigma$ is the noise multiplier that determines the privacy strength (more noise means more privacy and a lower $\epsilon$) and is determined based on the privacy budget ($\epsilon, \delta$), type of privacy accountant~\cite{dpsgd-Abadi2016,dp-gdp-koskela2022individual,dp-rdp-mironov,dp-prv-gopi} utilized to track the privacy loss $\epsilon$, number of gradient update steps $T$, and the sampling rate $q$ defined as $q=\frac{B}{|\mathcal{D}|}$ where $B$ is the expected batch size and $|\mathcal{D}|$ is the training dataset size. 
The term expected batch size is used here since DP-SGD does not use a fixed batch size in each training iteration but rather utilizes poisson sampling to generate a variable-size batch in each iteration.
Since DP-SGD adds noise before optimization, it can also be easily extended to work with other machine learning optimizers, such as Adam~\cite{kingma2017adam}. Refer to~\cref{app:dp-adam} for a complete pseudocode of the DP-SGD/Adam algorithm.

\subsubsection{Federated Learning}
\label{sec:prelim-fl}
In federated learning, a machine learning (ML) model is trained across multiple remote clients in a distributed manner, where the local data of each client remains onsite, thus preserving its privacy.
To train a ML model in a federated setting, McMahan \etal~\cite{fl-McMahan2016} introduced the well-known federated average (FedAvg) algorithm. Given $K$ number of total clients in a distributed setting, each round of FedAvg training involves sending model parameters from the server to a $C$ fraction of clients. 
Each client trains its own model replica on its local dataset for $E$ number of epochs, and then returns the updated model parameters to the server. Upon receiving model parameters from multiple clients, the server averages them and sends them back to the clients for the next training round. The complete pseudocode for the FedAvg algorithm is given in~\cref{app:fedavg}.
\section{Private Document KIE}
Our starting point for building private KIE systems was to leverage the power of large pretrained document foundation models. Large self-supervised pretrained models \cite{bert-devlin2019,kie-tilt-Powalski2021,kie-layoutlmv2-xu2021,kie-layoutlmv3-Huang2022} generally contain critical knowledge about the specific domain they are trained on, and have previously shown to greatly facilitate private training on downstream tasks~\cite{dp-Dupuy2022,dp-Li2021}. For these reasons, we built our private KIE model upon the state-of-the-art pretrained document transformer model \layoutlm{}.
\subsection{Model Configuration} 
\layoutlm{} is a unified text-image-layout transformer that learns cross-modal representations using self-attention~\cite{attention-vaswani2017,bert-devlin2019} and is pretrained on 11M document images from the large-scale IIT-CDIP public dataset~\cite{iit-cdip}. 
While a number of multimodal document foundation models have recently emerged~\cite{kie-tilt-Powalski2021,kie-layoutlmv2-xu2021,kie-formnet-Lee2022}, we found LayoutLMv3 to be the best choice  for two primary reasons: (1) the code and pretrained checkpoints of the model are publicly accessible, and (2) the model exhibits the best performance among all currently available multimodal transformers for the KIE task in a non-private setting.
LayoutLMv3 takes as an input a sequence of text embeddings $\mathbf{X_t}=\mathbf{x_t}_{1:L}$ and a sequence of image patch embeddings $\mathbf{X_I}=\mathbf{x_I}_{1:M}$, where $L$ and $M$ denote the lengths of the respective sequences, and $L+M$ represents the overall sequence length if both modalities are used. In addition to text and image modalities, LayoutLMv3 introduces document layout information through 2D positional embeddings, which are generated from the normalized bounding box coordinates $[x_1, y_1, x_2, y_2]$ of each word in the document image and are added to the text embeddings  $\mathbf{X_t}=\mathbf{x_t}_{1:L}$.
For a complete overview of the model architecture, refer to \cref{app:layoutlm}.

\subsection{Non-private Fine-tuning}
To fine-tune the model for the KIE task in a non-private setting, we use a classification head to predict the entity labels of the final embeddings of each token, and train the model for a total of $E$ epochs using cross-entropy loss and Adam optimizer~\cite{kingma2017adam}. Whereas, the target entity labels for the tokens are assigned using the standard BIO (Beginning-Inside-Outside) tagging. The input text sequence is pre-processed using Byte-Pair Encoding (BPE) and [CLS] and [SEP] tokens are added at the beginning and end of each text sequence. After tokenization, the entity labels are only assigned to the first sub-token of each word and the maximum text sequence length is set to $L=512$. The image inputs are resized to the dimensions $H\times W=224 \times 224$ and split into patches of sizes $16\times16$, resulting in a total image sequence length of $M=196$. 
Unless stated otherwise, the model weights are initialized using the publicly available pretrained checkpoint \textit{layoutlmv3-base}. For the complete model configuration used in our work, refer to~\cref{app:model-config}.

\subsection{Private Fine-tuning via DP-Adam}
To train the model with \textit{example-level privacy}, we directly fine-tune it for the KIE task with DP-Adam, a variant of DP-SGD that utilizes Adam~\cite{kingma2017adam} for model optimization. To track privacy loss, we use R\'enyi DP (RDP)~\cite{dp-rdp-mironov} accountant and compute the noise multiplier $\sigma$ for target privacy levels of $\epsilon\in\{8,20\}$ and $\delta=\frac{1}{|\mathcal{D}_{train}|}$ for a training set of size $|\mathcal{D}_{train}|$ for all experiments. 
Additionally, we report the converted $\epsilon$ using the Gaussian Accountant~\cite{dp-gdp-koskela2022individual}, and the Private Random Variable (PRV) Accountant~\cite{dp-prv-gopi} (see \cref{app:privacy_accounting} for details on different accountants).
For private training, we follow the same pre-processing procedure as non-private training, with the exception of sequence length $L$, which we experiment with in this scenario with values of $L=128$ and $L=512$.

\subsection{Private Fine-tuning via Federated Learning}
\label{exp:fl}
To investigate private training with federated learning (FL), we consider a scenario in which each client holds a small dataset of their own that contributes to the distributed training of a centralized model, as described in \cref{sec:prelim-fl}. This scenario is common in industrial document analysis pipelines in which a single service provider manages the data of several clients at the same time. We consider four different settings with total number of clients $K\in\{2,4,8,16\}$ and set the fraction of randomly sampled clients per round $C=0.5$ for settings with $K>2$. 
To train the models, we utilize the FedAVG algorithm with total number of FL rounds $T$ set to a value of  $\frac{E}{C}$,
where $E$ is the number of epochs that were used to fine-tune the model in the non-private setting.

\subsection{Private Federated Fine-tuning via FeAm-DP}
The starting point for developing our proposed FeAm-DP algorithm with \textit{example-level privacy} was to directly scale the standalone DP-SGD/Adam\cite{dpsgd-Abadi2016} algorithm to a global federated optimization setting similar to the work by Reddi \etal~\cite{fl-opt-reddi2021adaptive}.
While a few previous works have explored DP in a FL setting, they either deal with client-level privacy~\cite{fl-rnn-McMahan2017LearningDP} or apply the FedAVG algorithm in combination with DP-SGD~\cite{fl-mercier2021evaluating} to each client independently, providing no notion of global privacy. 
Our proposed algorithm mainly differs from these approaches in two ways:
\begin{itemize}
	\item[--] FeAm-DP provides \textit{example-level privacy} as opposed to client-level privacy introduced in ~\cite{fl-rnn-McMahan2017LearningDP}. 
	\item[--] Instead of defining privacy per client as done in FedAVG-DP~\cite{fl-mercier2021evaluating}, our proposed algorithm aims to simply upscale DP from a single client to multiple clients while keeping the same privacy constraints on a global level.
\end{itemize}
\cref{alg:dp-fedadam} provides the pseudocode for the proposed algorithm FeAm-DP. The algorithm is initialized with pretrained weights $\theta_o$, and $m$ number of clients to be sampled each round. 
In each training round, the server samples a subset of clients $\mathcal{S}_t$ and then performs a single update step on each client using the weights $\theta_{t-1}$ of the previous round. In the update step, each client samples a local batch $\mathcal{B}_{kt}$, computes the loss, clips the gradients according to the gradient norm $S$ and adds noise to the gradients. The noisy gradients are then returned to the server, averaged, and used for global optimization using the Adam optimizer~\cite{kingma2017adam}. 
Recall that in the standalone DP-SGD~\cite{dpsgd-Abadi2016} on a single client, the gradient update step is given as follows:
\begin{equation}
	\mathbf{\tilde{g}} = \frac{1}{q|\mathcal{D}|}(\sum_{i=1}^{q|\mathcal{D}|}\mathbf{\bar{g}}(x_i) + \mathcal{N}(0, \sigma^2 S^2\mathbf{I})
	\label{eq:g}
\end{equation}
Where $q$ is the per-example sampling rate, $\mathcal{D}$ is the training dataset, $\mathbf{\bar{g}}(x_i)$ is the per-example clipped gradient. Note that the term $q|\mathcal{D}|$ simply represents the expected batch size $B$.
In a FL environment, each client $k$ has its own small dataset $\mathcal{D}_k$ of size $|\mathcal{D}_k|$. We assume that each client has a dataset of size $\frac{|\mathcal{D}_{train}|}{K}$ with $K$ equal partitions of the original dataset $|\mathcal{D}_{train}|$. Given that only a fraction $C$ of these clients are sampled every round for training, the gradient averaging step after applying a single DP gradient update (as done in \cref{eq:g}) step per client can be written as follows:
\begin{align}
	\mathbf{\tilde{g}}_{fed} &= \frac{1}{CK} \sum_{k=1}^{CK} \frac{1}{q_k|\mathcal{D}_k|}(\sum_{i=1}^{q_k|\mathcal{D}_k|}\mathbf{\bar{g}}_k(x_i) + \mathcal{N}(0, \sigma_k^2 S^2\mathbf{I}))\\
	\mathbf{\tilde{g}}_{fed} &= \frac{1}{CKq_k\frac{|\mathcal{D}|}{K}}(\sum_{k=1}^{CK}\sum_{i=1}^{q_k\frac{|\mathcal{D}|}{K}}\mathbf{\bar{g}}_k(x_i) + \sum_{k=1}^{CK}\mathcal{N}(0, \sigma_k^2 S^2\mathbf{I}))\\
	\mathbf{\tilde{g}}_{fed} &= \frac{1}{\mathcolorbox{cyan}{Cq_k|\mathcal{D}|}}(\sum_{j=1}^{\mathcolorbox{cyan}{Cq_k|\mathcal{D}|}}\mathbf{\bar{g}}(x_j) + \mathcal{N}(0, \mathcolorbox{pink}{CK\sigma_k^2} S^2\mathbf{I})\label{eq:gfed-3})
\end{align}

\begin{figure}[bp]
	\centering	
	\begin{subfigure}{0.48\linewidth}
		\includegraphics[width=\linewidth]{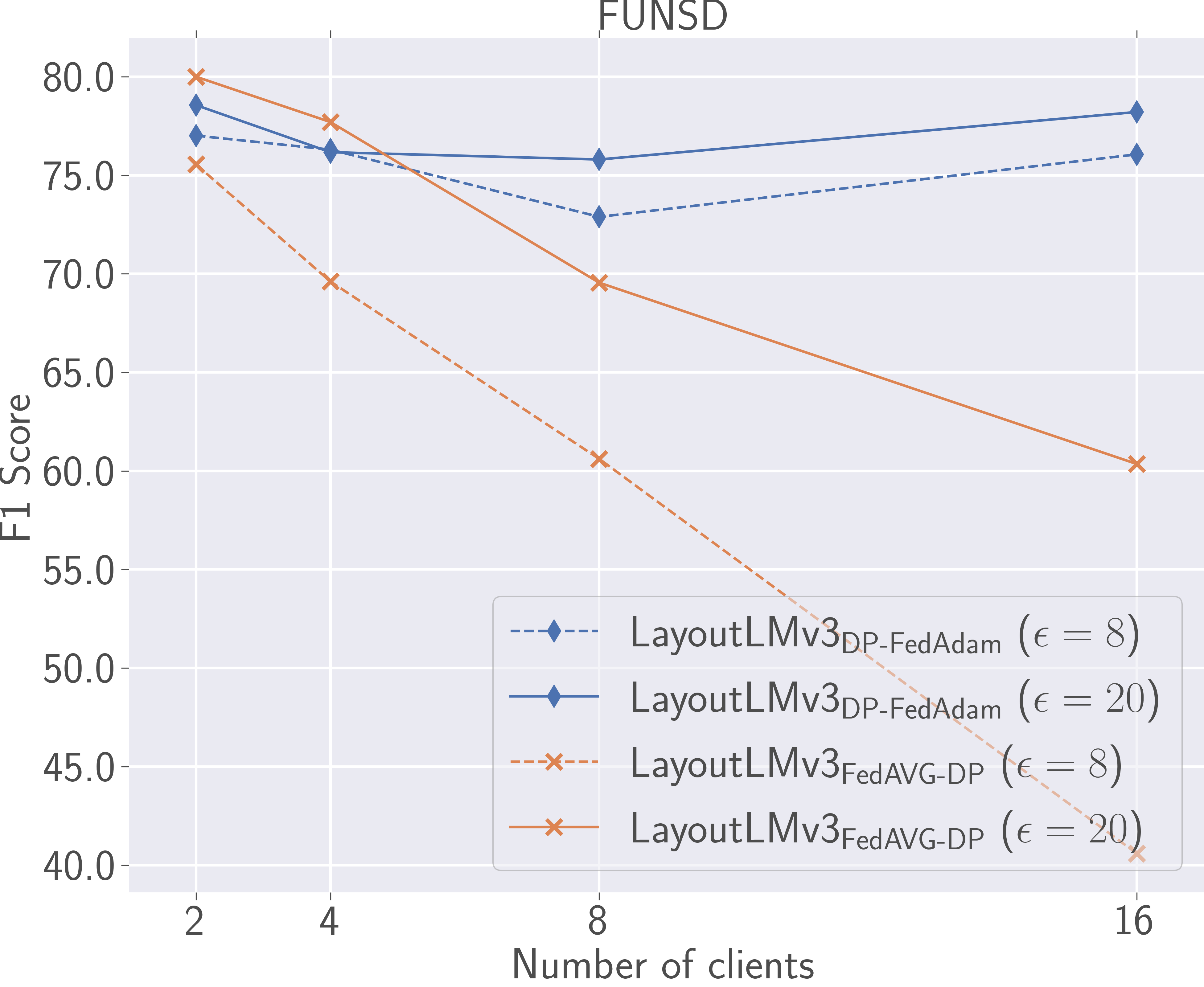}
	\end{subfigure}
	\hfill
	\begin{subfigure}{0.48\linewidth}
		\includegraphics[width=\linewidth]{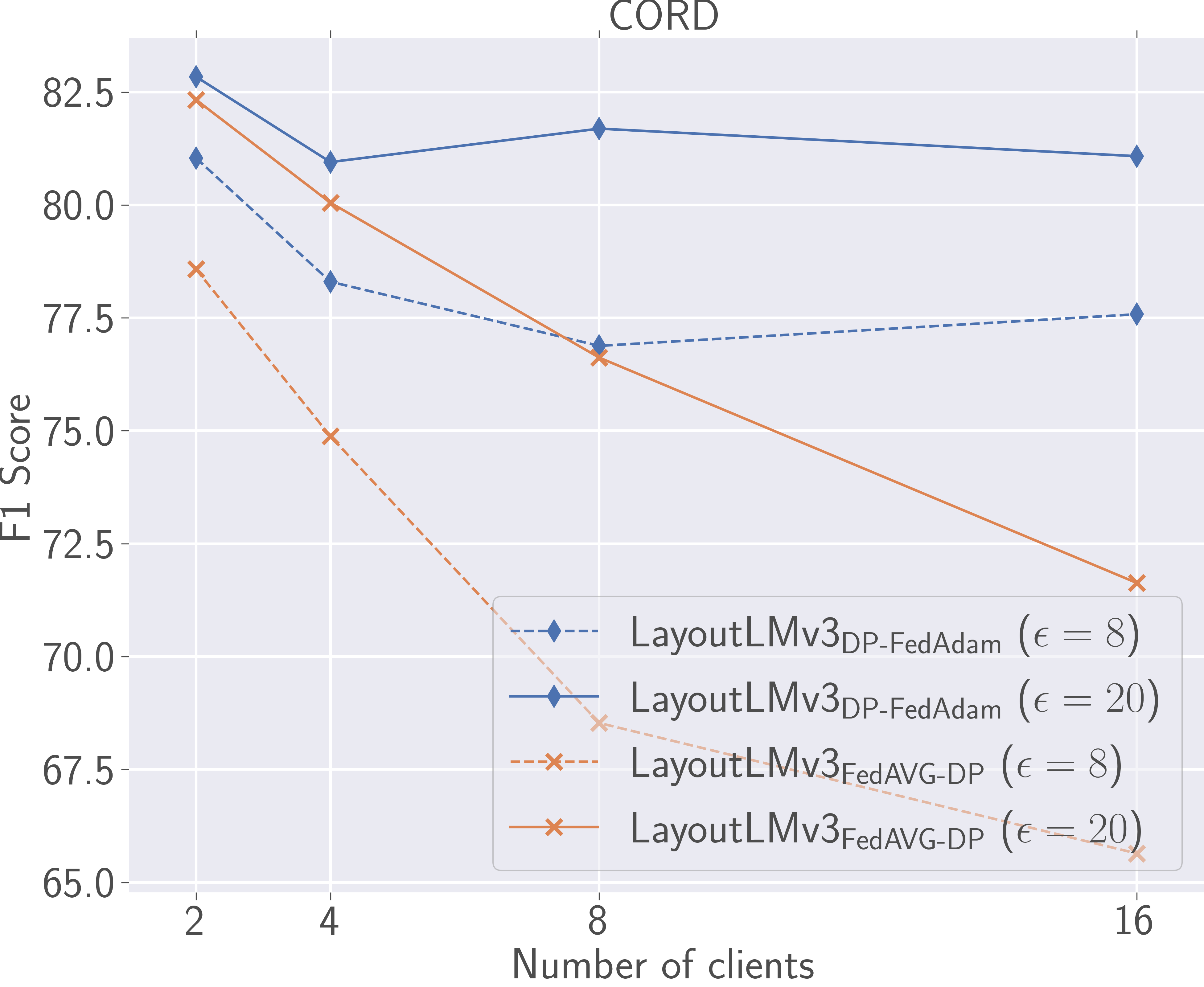}
	\end{subfigure}
	\caption{The performance of FedAVG-DP~\cite{fl-mercier2021evaluating} degrades substantially with increasing number of clients whereas our proposed FeAm-DP algorithm remains largely unaffected.}
	\label{fig:fedavg-dp-vs-dp-fedadam}
	\vspace{-1em}
\end{figure}

By comparing \cref{eq:gfed-3} with \cref{eq:g}, it becomes apparent that the only difference that arises when scaling DP from standalone scenario to multiple clients with gradient averaging is in the sampling rate (highlighted in blue) and the per-client noise multiplier (highlighted in red). From this it follows that to achieve the same sampling rate of $q$ in a federated setting, we must set the per-client sampling rate $q_k=\frac{q}{C}$. Similarly, to achieve the noise multiplier of $\sigma$ in a federated setting, we must set the per-client noise multiplier as $\sigma_k=\frac{\sigma}{\sqrt{CK}}=\frac{\sigma}{\sqrt{m}}$.
In addition to the sampling rate $q$, in standalone DP setting~\cite{dpsgd-Abadi2016}, the total privacy cost  $(\epsilon,\delta)$ spent during the training process also depends on the the total number of gradient update steps $U$. For standalone DP, the number of gradient update steps can be simply computed as $U=\frac{1}{q}E$, and $E$ is the number of training epochs. 
Since in our proposed algorithm, each client only performs a single update step on the model, the total number of FL communication rounds can be simply set as $T=U=\frac{1}{q}E$. 

It is worth mentioning that in addition to providing a global privacy perspective, our proposed algorithm provides some major advantages over the FedAVG-DP algorithm~\cite{fl-mercier2021evaluating} under federated setting with small datasets. In FedAVG-DP~\cite{fl-mercier2021evaluating}, each client updates its own model replica with noisy gradients on a local batch of samples for $E$ number of epochs. 
Since the choice of batch size is limited by privacy constraints $(\epsilon,\delta)$, as the number of clients increases and the local datasets $|D_k|$ become smaller, the expected batch size (since $B=q|D_k|$) for model optimization per client for a fixed privacy budget $\epsilon$ also reduces. 
\SetKwProg{ClientUpdate}{ClientUpdate}{\textbf{:}}{\KwRet{$\mathbf{\tilde{g}}_{k,t}$}}
\begin{algorithm}[!t]\footnotesize\SetAlgoLined
	\caption{FeAm-DP}\label{alg:dp-fedadam}
	\KwIn{Learning rate $\eta$, gradient clipping norm $S$, sampling rate $q$, target $(\epsilon,\delta)$, total clients K, clients sampling rate $C$, privacy accountant $\mathcal{M}$, total FL rounds $T$}
\Server{
	\KwInit{$\theta_0$, $m \gets CK$, $\sigma \gets \mathcal{M}.\text{get\_noise\_multiplier}(q,\epsilon,T)$}
	\For{each round $t=1,\dots,T$}{
		$\mathcal{S}_t \gets (\text{sample a set of $m$ clients from $K$})$\\
		\ForEachParallel{each client $k\in\mathcal{S}$}{
			$\sigma_k \gets \frac{\sigma}{\sqrt{m}}$\\
			$\mathbf{\tilde{g}}_{k,t} \gets \text{ClientUpdate}(k, \theta_{t-1}, \sigma_{k}, S, C)$
		}
		$\mathbf{\tilde{g}}_{t} \gets \sum_{k\in\mathcal{S}_t}\frac{n_k}{n}\mathbf{\tilde{g}}_{k,t}$\\	
		$\theta_{t+1} \gets$ \textbf{GlobalAdamUpdate}$(\eta, \theta_t, \mathbf{\tilde{g}}_{t})$\\
		print  $\mathcal{M}.\text{get\_privacy\_spent}(q,\sigma,t,\delta)$
	}
}	
\ClientUpdate{($k, \theta_t, \sigma_{k}, S, C$)}{
	\KwIn{$\mathcal{L}(\theta) = \frac{1}{B}\sum_i\mathcal{L}(\theta, x_i)$, $\mathcal{D}_k$ of size $|\mathcal{D}_k|$}
	$\mathcal{B}_{kt} \gets ($sample a batch of size $B_k$ with sampling probability $q_k = \frac{q}{C}$)\\
	\ForEachGradient{$x_i\in\mathcal{B}_{kt}$}
	{$\mathbf{g}_{k,t}(x_i) \gets \nabla_{\theta_{k,t}} \mathcal{L}(\theta_{k,t}, x_i)$\\
		$\mathbf{\bar{g}}_{k,t}(x_i) \gets \mathbf{g}_{k,t}(x_i) / max(1, \frac{||\mathbf{\bar{g}}_{k,t}(x_i)||_2}{S})$}
	$\mathbf{\tilde{g}}_{k,t} \gets \frac{1}{B_k}(\sum_i\mathbf{\bar{g}}_{k,t}(x_i) + \mathcal{N}(0, \sigma{k}^2S^2\mathbf{I})$\\
}
\end{algorithm}
\begin{table}[b]
\centering
\begin{center}
	\resizebox{0.9\linewidth}{!}{
		\begin{tabular}{@{}lcccccc@{}}
			\toprule
			Dataset & Train & Train $(L=512)$ & Train $(L=128)$ &Validation & Test & Entity Labels\\
			\midrule
			FUNSD~\cite{funsd-jaume2019} & $149$ &$149$& $389$ &- & $50$&$4$\\
			CORD~\cite{cord-park2019} & $800$ &$800$& $840$ & $100$ & $100$&$31$\\
			SROIE~\cite{sroie-Huang2019} & $626$ &$630$& $1601$ &$-$ & $347$&$5$\\
			WildReciepts~\cite{wildreceipts-sun2021} & $1267$ &$1267$&$2204$ & $-$ & $472$ &$25$\\
			DOCILE~\cite{docile} & $5180$ &$9768$&$29406$& $500$ & $1000$ & $37$\\
			XFUND~\cite{xfund} & $149$ &$188$& $562$ & $-$ & $50$ & $4$\\
			\bottomrule
		\end{tabular}
	}
\end{center}
\vspace{-1em}
\caption{An overview of the datasets is provided.}
\label{tab:datasets}
\vspace{-1em}
\end{table}
This in turn degrades the performance of FedAVG-DP~\cite{fl-mercier2021evaluating} as the underlying application of DP-SGD/Adam per client is strongly influenced by batch size. \cref{fig:fedavg-dp-vs-dp-fedadam} illustrates this problem by comparing the performance of FedAVG-DP~\cite{fl-mercier2021evaluating} and our proposed FeAm-DP under different numbers of clients. As can be seen, the performance of the FedAVG-DP~\cite{fl-mercier2021evaluating} algorithm is significantly diminished as the number of clients increases whereas our proposed algorithm remains largely unaffected.

Overall, our proposed algorithm offers privacy on a global level while keeping the local client data onsite. Furthermore, because noise is added on the client end, the client gradients also remain protected under DP during each FL communication round.
For our experiments with FeAm-DP, similar to FL settings, we consider four different settings of clients with $K\in\{2,4,8,16\}$ and $C=0.5$ for settings with $K>2$. 

\section{Experiments}
\subsection{Datasets}
\begin{figure}[b]
	\centering	
	\includegraphics[width=0.6\linewidth]{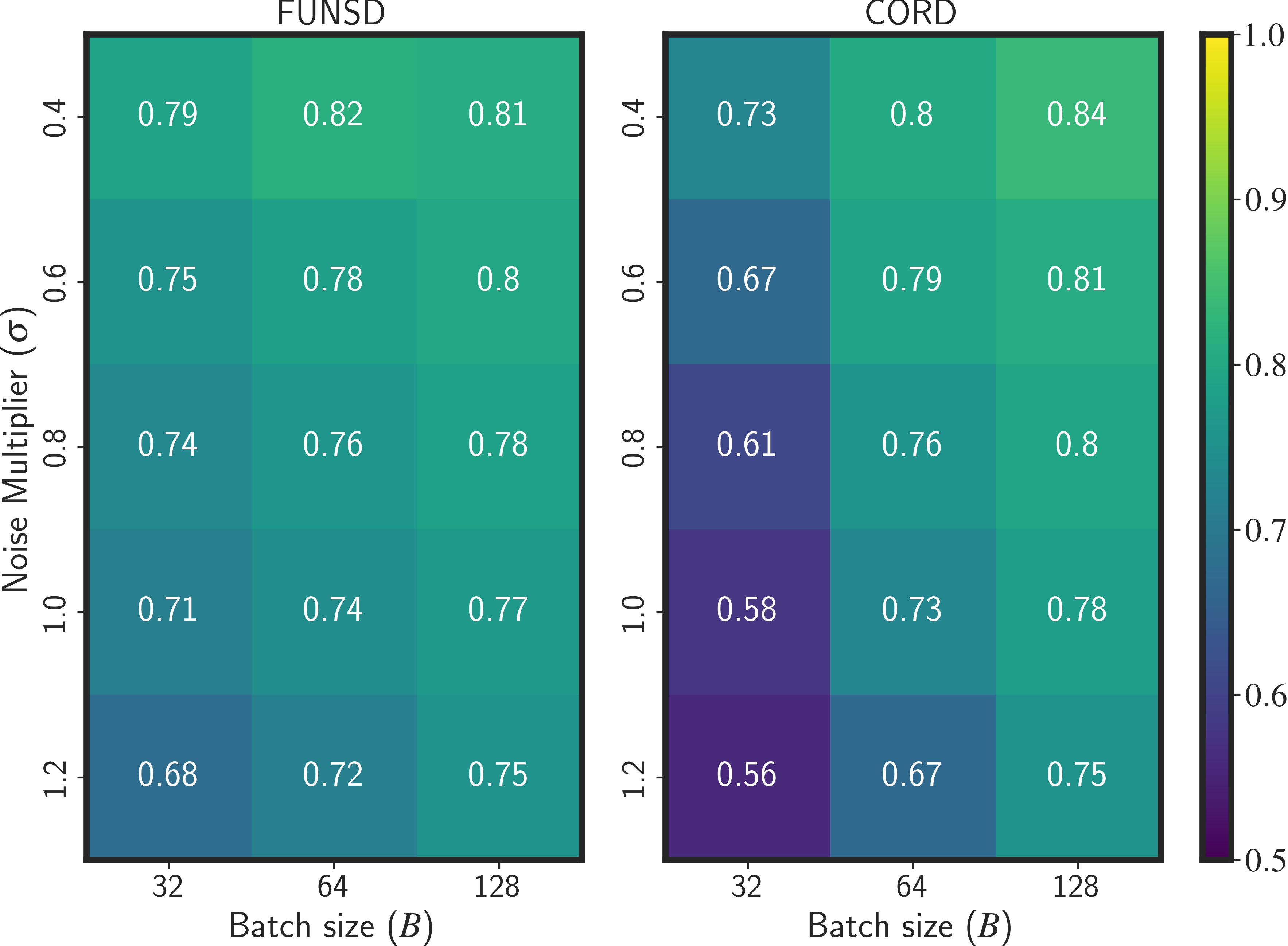}
	\caption{Larger batch sizes $B$ result in better performances over a range of noise values $\sigma$. Results reported here were obtained with $L=512$, and $\eta=1.0e{-}3$.}
	\label{fig:nm_vs_bs}
	\vspace{-1em}
\end{figure}
\label{sec:datasets}
For a thorough analysis of our private training approaches, we conduct experiments on a variety of datasets as shown in~\cref{tab:datasets}. 
FUNSD~\cite{funsd-jaume2019} is a form-understanding dataset which contains $199$ documents with $4$ different semantic entity types. 
SROIE~\cite{sroie-Huang2019} is receipts understanding dataset with $626$ training samples and $347$ test samples with $5$ target entity types. 
CORD~\cite{cord-park2019} is another receipt understanding dataset containing $1000$ and $31$ different target entities.
WildReciepts~\cite{wildreceipts-sun2021} is a relatively large receipts dataset which includes $1267$ and $472$ training and test samples and consists of $25$ target entities. 
DOCILE~\cite{docile} is a recently released large-scale document KIE dataset containing $6206$ training samples, $500$ validation images, and $1000$ test images, and about ${\sim}100$k synthetically generated samples with a total of $37$ target entities. Since DOCILE~\cite{docile} test set annotations are not made public by the authors, we only report model performance on the validation set. 
XFUND~\cite{xfund} is a multilingual form understanding benchmark dataset that includes human-labeled forms with key-value pairs in $7$ languages. From this dataset, we perform experiments on the Chinese language subset which contains $149$ training images and $50$ test images.
Note that for training LayoutLMv3 on XFUND dataset, we use the Chinese language pretraining checkpoint namely \textit{layoutlmv3-base-chinese}. For additional dataset details, see \cref{app:dataset_details}.

\begin{figure}[t]
	\centering	
	\includegraphics[width=0.8\linewidth]{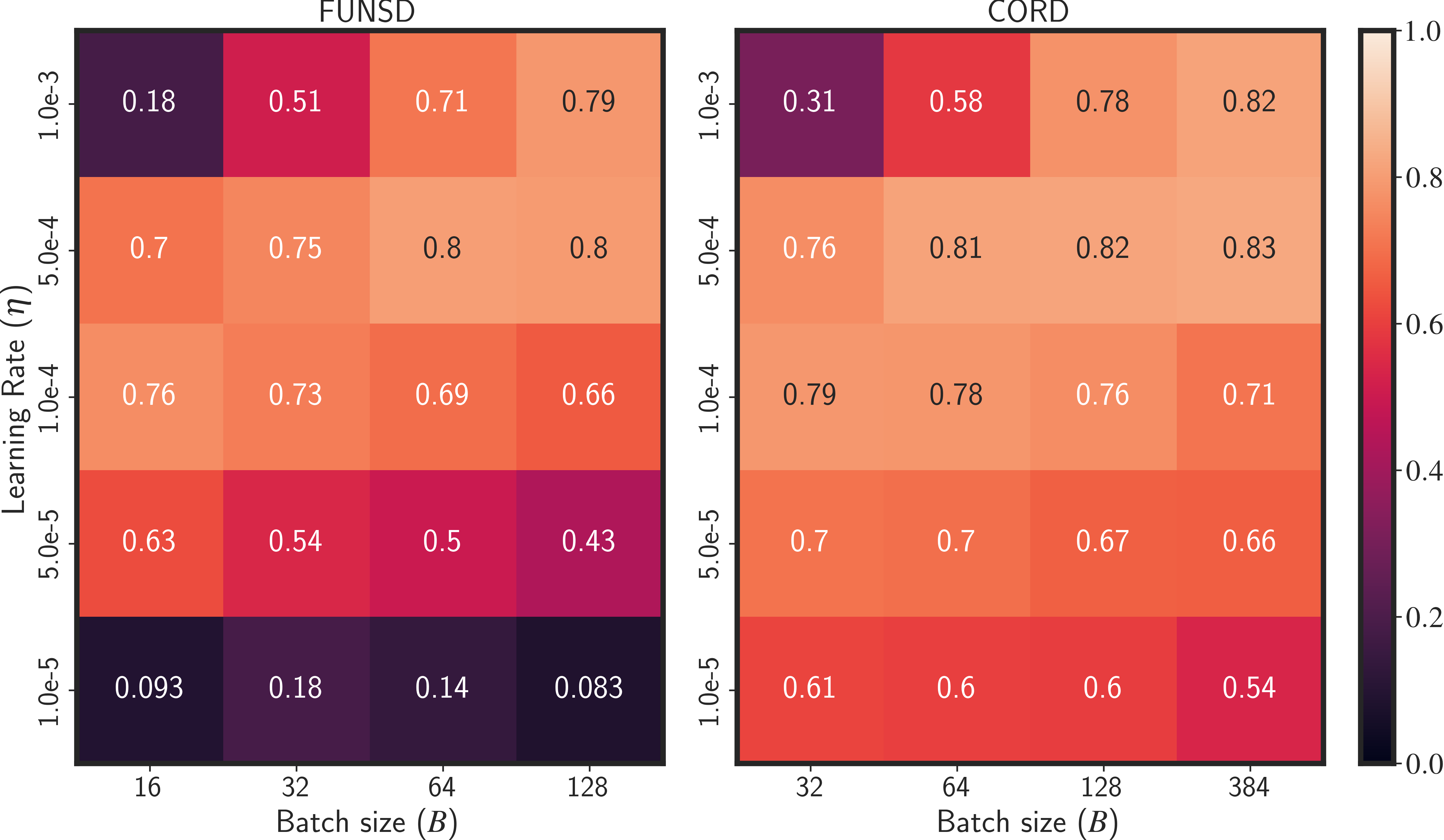}
	\caption{Higher learning rates $\eta$ and larger batch sizes $B$ result in better performances. Results reported here were obtained with $\epsilon=20$, and $L=128$.}
	\label{fig:lr_vs_bs}
	\vspace{-1em}
\end{figure}
\subsection{Guidelines for Effective DP Fine-Tuning}
DP-Adam~\cite{dpsgd-Abadi2016} algorithm is known to be particularly sensitive to the choice of hyperparameters~\cite{dp-Li2021,dp-papernot2020making,dp-Dupuy2022}.
In this section, we thoroughly investigate the impact of various training parameters on the performance of DP fine-tuning. Subsequently, we present guidelines based on our observations to attain an optimal trade-off between privacy and utility. 
\label{sec:dp-hyper-param-tuning}
\begin{figure}[b]
	\centering	
	\includegraphics[width=0.8\linewidth]{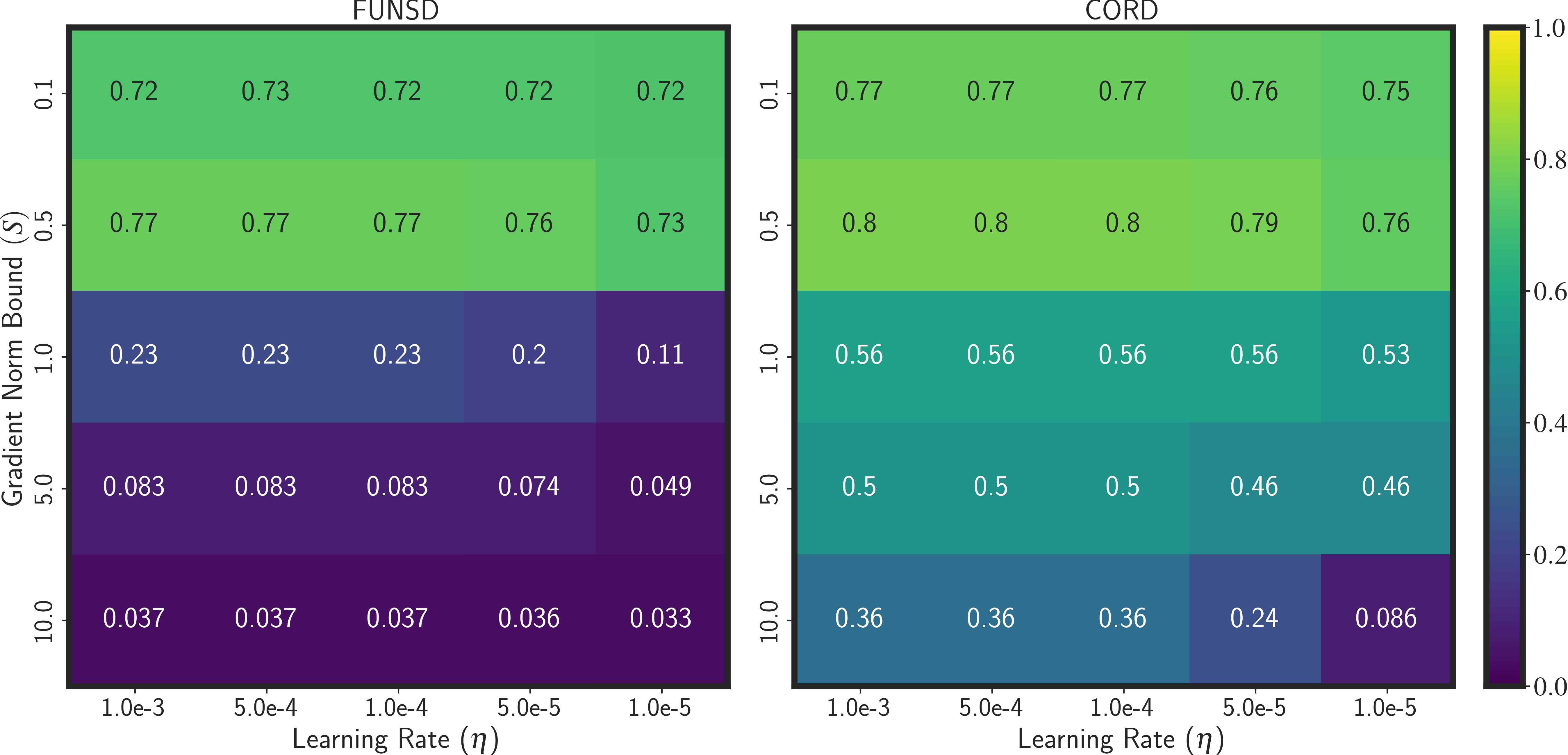}
	\caption{Lower values of clipping norm $S$ result in better performances for a range of learning rates $\eta$. Results reported here were obtained with $\epsilon=20$, and $L=128$.}
	\label{fig:lr_vs_cn}
	\vspace{-1em}
\end{figure}
\subsubsection{Batch size, Learning Rate, and Clipping Norm}
Our experiments indicate that the training parameters batch size, learning rate, and clipping norm were all equally important and considerably affected the model performance under DP-Adam. 
\cref{fig:nm_vs_bs} illustrates the effects of increasing batch size $B$ on the model performance under different values of the noise multiplier $\sigma$. It can be observed that across low to high noise regimes, larger batch sizes always resulted in better performances. 
Generally we observed that batch sizes corresponding to sampling rate $q$ in the range $(\frac{1}{10}, \frac{1}{3})$ performed the best and batch sizes corresponding to sampling rates above $\frac{1}{3}$ did not show a significant positive improvement in the model performance. 
Similarly, it can be seen from \cref{fig:lr_vs_bs} that across varying batch sizes, higher learning rates resulted in improved performances. However, we observed that too high learning rates were also unsatisfactory, resulting in training instability. For our experiments, we found a learning rates between the range of $\eta\in\{1.0e{-}3, 5.0e{-}4\}$ to work out the best across all datasets.
The effect of clipping norm $S$ on model performance was also very prominent. As can be observed from \cref{fig:lr_vs_cn}, for a fixed batch size, smaller values of $S$ resulted in considerably better performances. For our experiments, we adjusted the clipping norm to a small value of $S=0.1$.
\begin{guidelines}
To achieve an optimal privacy-utility tradeoff for fine-tuning DP-Adam for KIE with a fixed number of epochs $E$, we recommend choosing large batch sizes, relatively large learning rates compared to non-private fine-tuning, and small values of gradient clipping norm $S$.
\end{guidelines}

\begin{table}[t]
	\centering
	\begin{center}
		\resizebox{0.7\linewidth}{!}{
			\begin{tabular}{@{}llccccc@{}}
				\toprule
				Dataset&Method&$\epsilon=8$&$\epsilon=20$\\
				\midrule\multirow{5}{*}{FUNSD}
				& \layoutlmbase                         &68.81&75.27\\
				& \layoutlmbasetsft{SROIE} 				&66.32 $\downarrowc$&73.31 $\downarrowc$\\
				& \layoutlmbasetsft{WildReceipts} 		&63.81 $\downarrowc$&74.96 $\downarrowc$\\
				& \layoutlmbasetsft{DOCILE-Synthetic} 		&65.09$\downarrowc$&72.27$\downarrowc$\\
				& \layoutlmbasetsft{CORD} 				&72.49 $\uparrowc$&79.19 $\uparrowc$\\
				\midrule\multirow{4}{*}{CORD}
				& \layoutlmbase                         &72.08&78.28\\
				& \layoutlmbasetsft{SROIE} 				&70.11 $\downarrowc$&77.41 $\downarrowc$\\
				& \layoutlmbasetsft{DOCILE-Synthetic} 		&70.19$\downarrowc$&76.39$\downarrowc$\\
				& \layoutlmbasetsft{WildReceipts} 		&75.52 $\uparrowc$&79.53 $\uparrowc$\\
				\midrule\multirow{4}{*}{SROIE}
				& \layoutlmbase                         &69.79&77.05\\
				& \layoutlmbasetsft{CORD} 				&71.13 $\uparrowc$&75.36 $\downarrowc$\\
				& \layoutlmbasetsft{WildReceipts}		&73.20 $\uparrowc$&77.69 $\uparrowc$\\
				& \layoutlmbasetsft{DOCILE-Synthetic} 		&76.17$\uparrowc$&79.84$\uparrowc$\\
				\midrule\multirow{4}{*}{WildReceipts}
				& \layoutlmbase       					&71.26&78.44\\
				& \layoutlmbasetsft{SROIE} 				&69.15 $\downarrowc$&76.33 $\downarrowc$\\
				& \layoutlmbasetsft{DOCILE-Synthetic} 		&71.23$\downarrowc$&78.20$\downarrowc$\\
				& \layoutlmbasetsft{CORD} 				&73.38 $\uparrowc$&79.71 $\uparrowc$\\
				\bottomrule
			\end{tabular}
		}
	\end{center}
	\caption{Performance comparison of task-specific pretraining (TSP) for four datasets under different privacy budgets. Results reported here were obtained with $L=512$.}
	\label{table:tsft}
	\vspace{-1em}
\end{table}

\subsubsection{Task-specific Pretraining (TSP)}
To provide task-specific information to the model, we explored task-specific pretraining (TSP) which involved first training the model on other datasets considering them as public datasets, and then fine-tuning it on the target dataset under private settings.
To analyze the effects of TSP on model performance, we performed experiments on four datasets FUNSD, CORD, SROIE, and WildReceipts for which the results are given in \cref{table:tsft}. 
The term \layoutlmbasetsft{A} indicates that the model was first fine-tuned with a non-private setting on dataset A and then further fine-tuned under a private setting on the target dataset. 
Overall, the results indicate that with appropriate choice of datasets, TSP can lead to significant performance improvements ranging from ${\sim}2\%$ to ${\sim}6.5\%$. 
Another interesting observation from these results that may seem counterintuitive at first glance is that the overall impact of TSP on model performance was disparate, i.e. for a particular private dataset, only a select few datasets resulted in performance improvements. 
However, on closer examination we discovered that the performance improvements based on TSP were closely linked to the overlap or similarity between semantic entity labels of the fine-tuning dataset and the target dataset. 
For example, the CORD and WildReceipts datasets performed significantly better when used for TSP for each other since their entity labels shared a considerable amount of similarity. 
Similarly, as SROIE had complete entity overlap with both WildReceipts and DOCILE-Synthetic, it benefited from both of these datasets, in contrast to CORD, with which it only had limited entity overlap.

\begin{guidelines}
To maximize performance when using TSP in conjunction with DP for the KIE task, we recommend using public datasets whose target entity types are similar to those in the private dataset.
\end{guidelines}

\begin{figure}[t]
	\centering
	\begin{subfigure}{0.48\linewidth}
		\includegraphics[width=\linewidth]{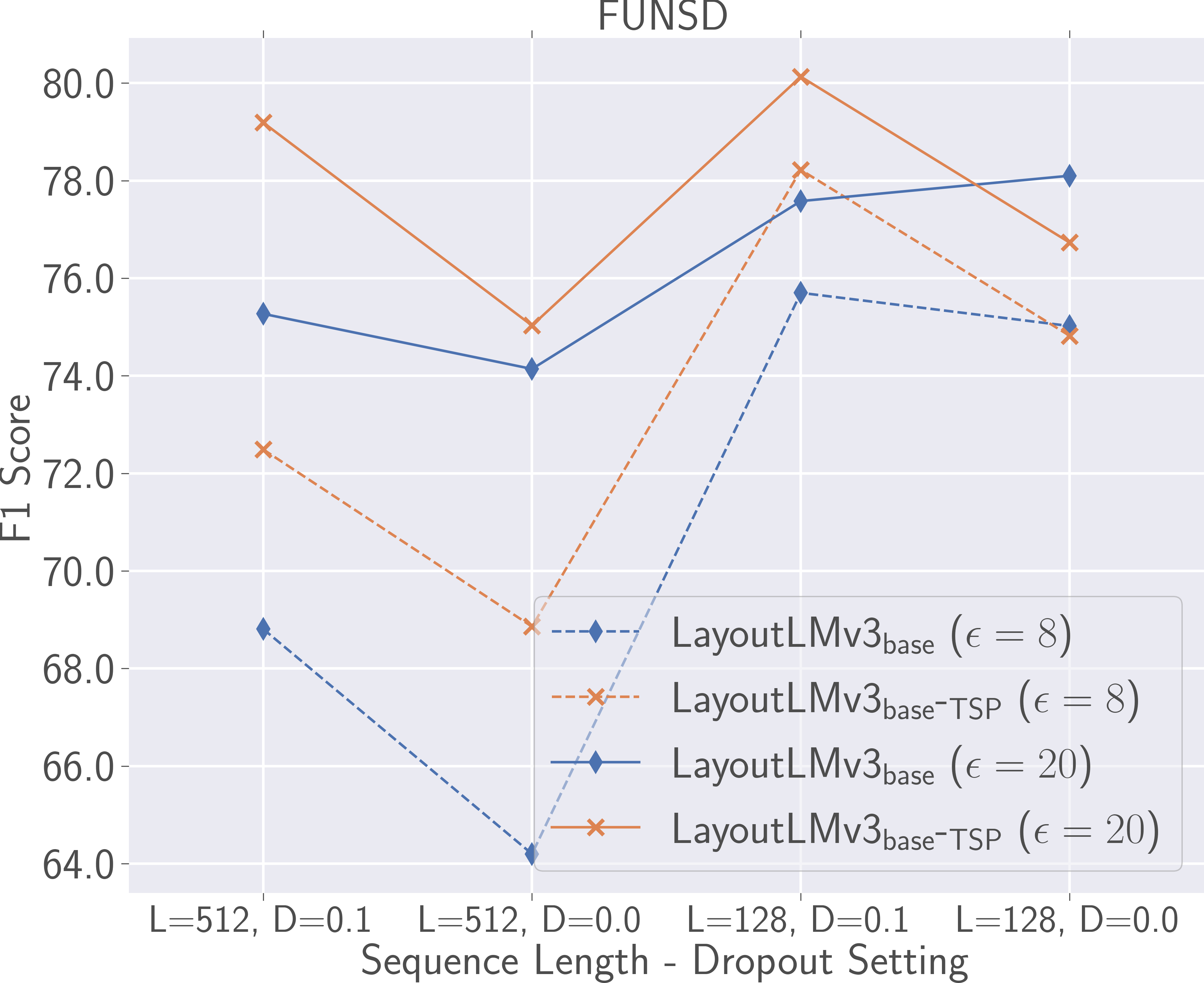}
	\end{subfigure}
	\hfill
	\begin{subfigure}{0.48\linewidth}
		\includegraphics[width=\linewidth]{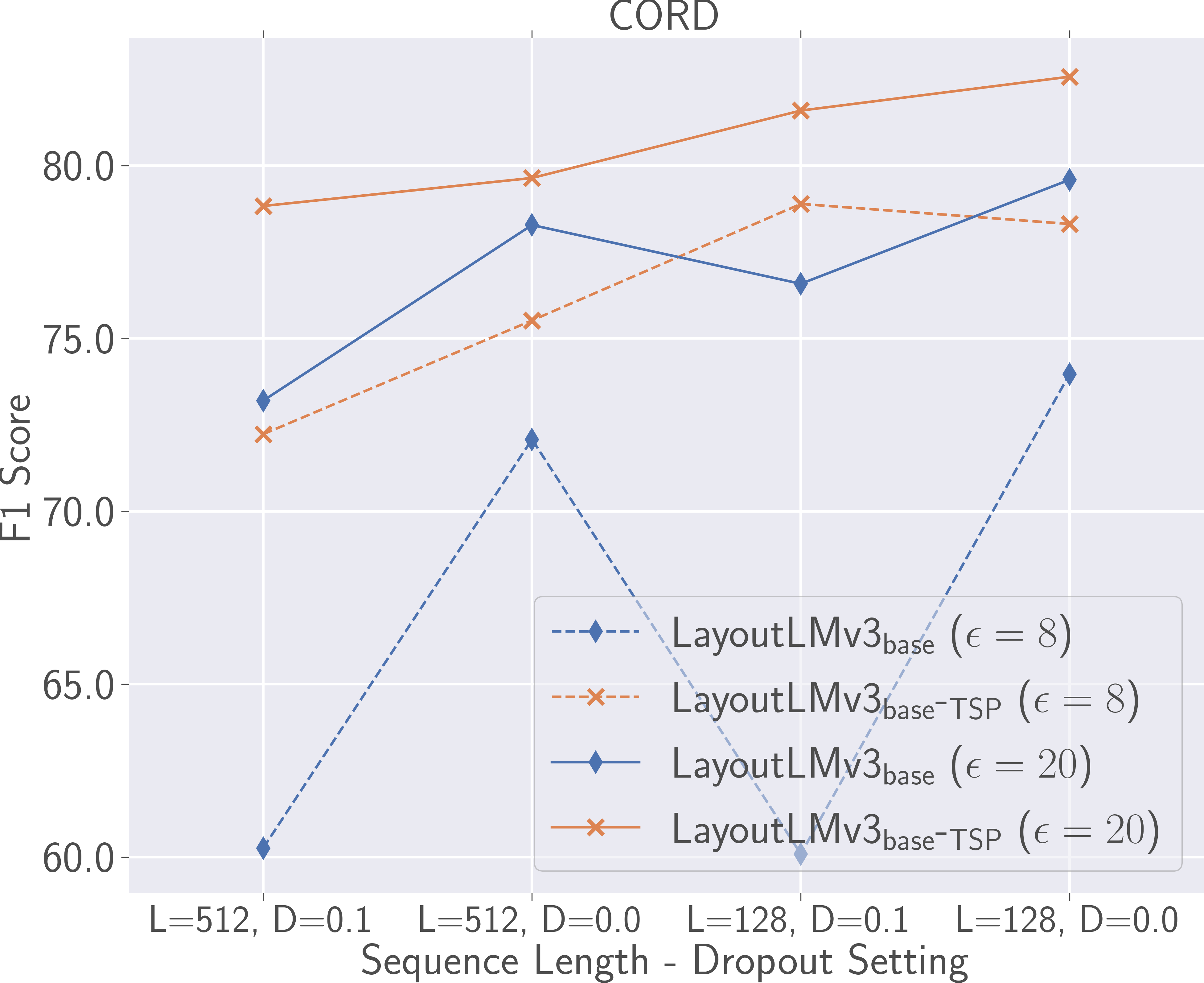}
	\end{subfigure}
	\caption{Sequence lengths of $L=128$ result in significantly better performances (especially under task-specific pretraining), whereas dropout shows a disparate effect.}
	\label{fig:dropout_seq_len}
	\vspace{-1em}
\end{figure}

\subsubsection{Sequence Length and Dropout}
For training on sequential data, sequence length is a crucial parameter, and an appropriate sequence length is necessary to provide the model with relevant context. A longer sequence length indicates a larger context but a smaller overall dataset size, while a shorter sequence length indicates the opposite. Since DP-SGD/Adam depends strongly on dataset size, we experimented with sequence lengths of $L\in\{128,512\}$ to see how it affects overall performance. Dropout is another significant hyperparameter that directly affects model performance. In our experiments, we experimented with dropout rates of $\{0.0, 0.1\}$. The performance of the model under varying settings of sequence length $L$, dropout and privacy budget $\epsilon$ on two datasets FUNSD and CORD is shown in \cref{fig:dropout_seq_len}. Overall, we observed that for most datasets, a sequence length of $L=128$ resulted in substantial performance boosts compared to $L=512$. However, on DOCILE a sequence length of $L=512$ performed better, likely due to the fact that the DOCILE dataset contains much larger text sequences (with $L\gg512$) per document than other datasets. Unlike sequence length, dropout showed a disparate effect on model performance. For smaller datasets FUNSD, XFUND, dropout${=}0.1$ generally resulted in better performances (especially in case of task-specific pretraining) whereas for the remaining datasets which were much larger in size, adding dropout considerably degraded the model performance. 

\begin{guidelines}
Based on above observations, we recommend choosing a smaller sequence length $L=128$ for datasets where average sequence sizes per document are also small ($\ll 512$), and choosing a large sequence length $L=512$ when average sequence sizes per document are also large ($\gg 512$). In addition, we recommend using no dropout for medium to large sized datasets and only to explore dropout values of $0.1$ for very small datasets.
\end{guidelines}

\begin{figure}[t]
	\centering
	\begin{subfigure}{0.48\linewidth}
		\includegraphics[width=\linewidth]{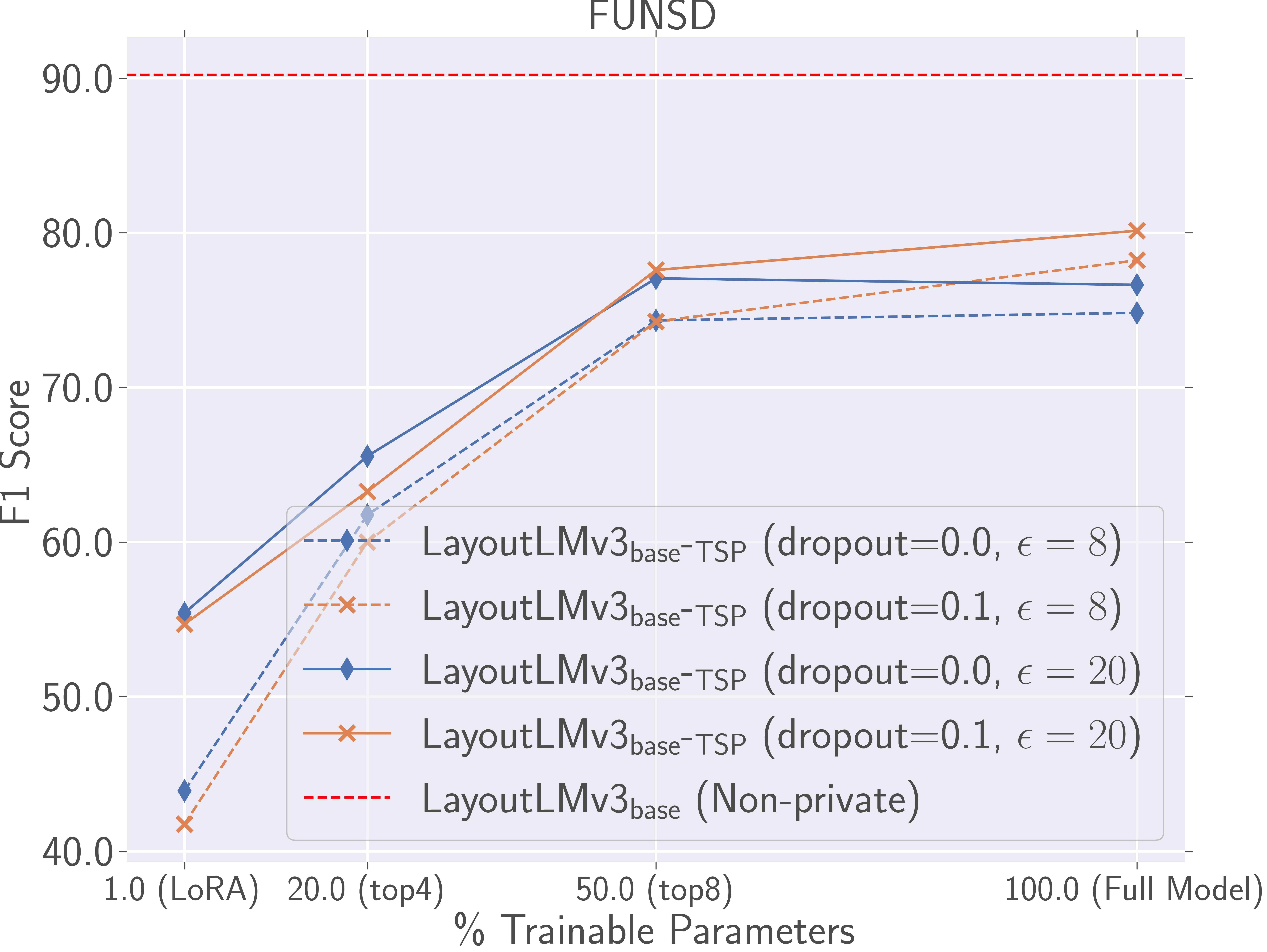}
	\end{subfigure}
	\hfill
	\begin{subfigure}{0.48\linewidth}
		\includegraphics[width=\linewidth]{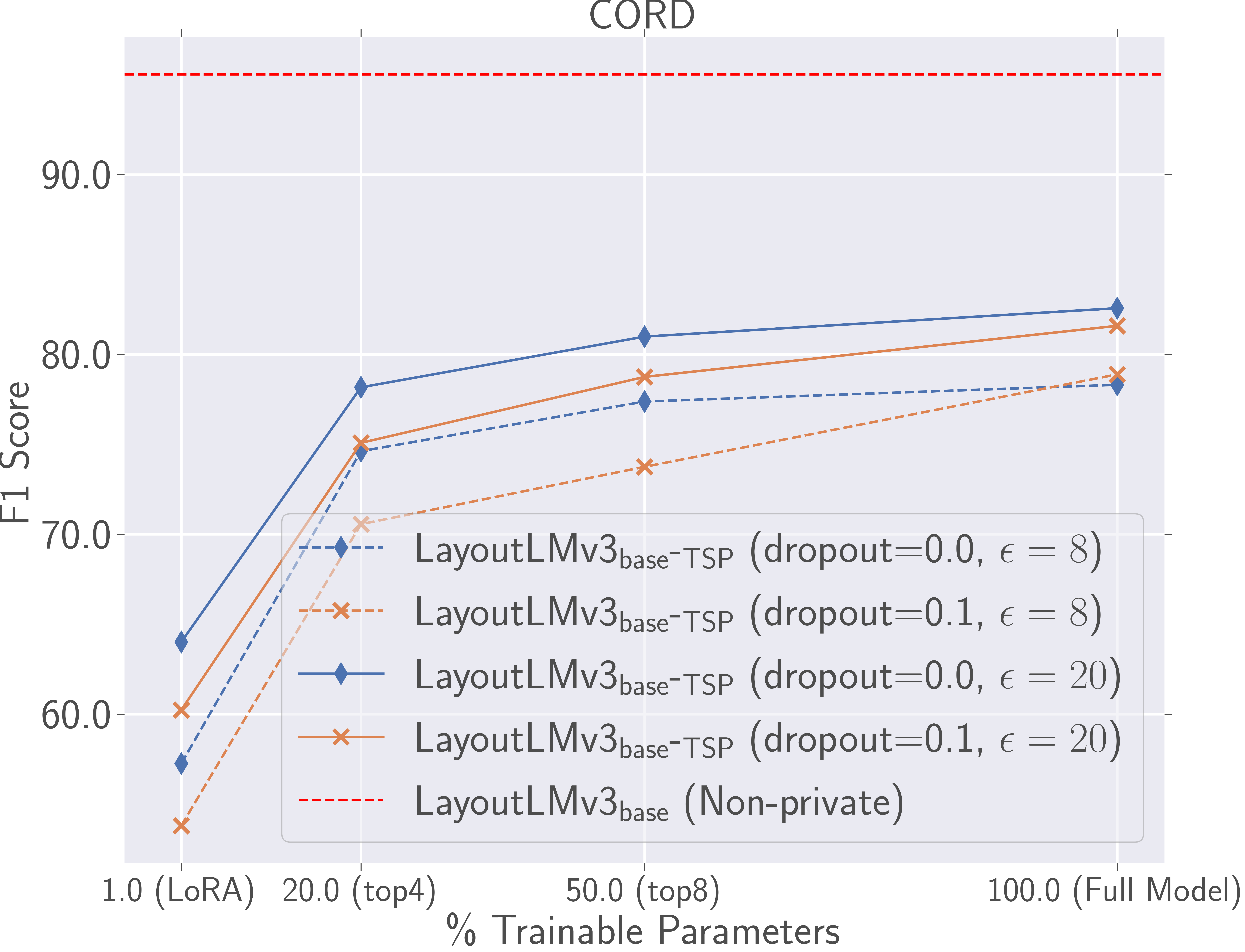}
	\end{subfigure}
	\hfill
	\caption{Full fine-tuning performs the best whereas top8 with only 50\% parameters results in performances similar to the full fine-tuning.}
	\label{fig:param-red}
	
	\vspace{-1.5em}
\end{figure}
\begin{table*}[!t]
	\centering
	
	\begin{center}
		\resizebox{0.95\linewidth}{!}{
			\begin{tabular}{@{}llccccccccccccccccccc@{}}
				\toprule
				\multirow{2.5}{*}{Dataset}&\multirow{2.5}{*}{Method}&\multirow{2.5}{*}{\scorenp}&\multicolumn{2}{c}{$\epsilon_{\text{G}}$}&\multicolumn{2}{c}{$\epsilon_{\text{PRV}}$}&\multicolumn{2}{c}{\scoredp}&\multicolumn{4}{c}{\scorefl}&\multicolumn{4}{c}{\scorefldp $(\epsilon=8)$}&\multicolumn{4}{c}{\scorefldp $(\epsilon=20)$}\\
				\cmidrule(lr){4-5}\cmidrule(lr){6-7}\cmidrule(lr){8-9}\cmidrule(lr){10-13}\cmidrule(lr){14-17}\cmidrule(lr){18-21}
				&&&$\epsilon=8$&$\epsilon=20$&$\epsilon=8$&$\epsilon=20$&$\epsilon=8$&$\epsilon=20$&K=2&K=4&K=8&K=16&K=2&K=4&K=8&K=16&K=2&K=4&K=8&K=16\\
				\midrule\multirow{3}{*}{FUNSD~\cite{funsd-jaume2019}}
				& \robertabase{} &65.62&\multirow{3}{*}{$\approx6.96$}&\multirow{3}{*}{$\approx19.52$}&\multirow{3}{*}{$\approx7.01$}&\multirow{3}{*}{$\approx20$}&33.93&39.42&64.14&64.08&57.09&46.59&29.07&33.58&34.24&33.79&38.88&38.32&35.26&39.59\\
				& \layoutlmbase{} (w/o Image) &87.20&&&&&65.66&71.68&86.12&86.55&83.84&82.18&67.90&63.82&67.31&70.29&73.59&70.72&74.61&74.88\\
				& \layoutlmbase{} &\textbf{90.22}&&&&&\textbf{78.22}&\textbf{80.12}&\textbf{88.48}&\textbf{87.73}&\textbf{88.94}&\textbf{84.40}&\textbf{77.01}&\textbf{76.30}&\textbf{72.89}&\textbf{76.06}&\textbf{78.56}&\textbf{76.17}&\textbf{75.80}&\textbf{78.21}\\ %
				\midrule\multirow{3}{*}{CORD~\cite{cord-park2019}}
				& \robertabase{} &93.50&\multirow{3}{*}{$\approx7.07$}&\multirow{3}{*}{$\approx19.35$}&\multirow{3}{*}{$\approx7.11$}&\multirow{3}{*}{$\approx17.94$}&72.38&75.61&93.20&93.88&91.58&88.23&68.89&73.31&71.44&69.12&76.39&76.34&75.28&74.92\\
				& \layoutlmbase{} (w/o Image) &\textbf{95.63}&&&&&77.39&80.43&96.07&95.32&\textbf{94.53}&87.27&80.71&74.19&75.32&74.52&80.77&80.83&80.10&79.76\\
				& \layoutlmbase{} &95.58&&&&&\textbf{78.30}&\textbf{82.56}&\textbf{96.34}&\textbf{95.63}&94.32&\textbf{87.86}&\textbf{81.04}&\textbf{78.30}&\textbf{76.88}&\textbf{77.58}&\textbf{82.84}&\textbf{80.95}&\textbf{81.69}&\textbf{81.08}\\
				\midrule\multirow{3}{*}{SROIE~\cite{sroie-Huang2019}}
				& \robertabase{} &91.01&\multirow{3}{*}{$\approx7.03$}&\multirow{3}{*}{$\approx19.04$}&\multirow{3}{*}{$\approx7.15$}&\multirow{3}{*}{$\approx17.95$}&70.41&75.36&89.50&88.92&88.95&86.11
				&74.35&71.30&70.96&66.37&77.50&76.72&75.59&75.18\\
				& \layoutlmbase{} (w/o Image) &90.60&&&&&74.19&76.56&90.06&90.65&89.82&87.42&
				75.39&73.73&72.98&72.10&79.40&77.71&78.40&77.69\\
				& \layoutlmbase{} &\textbf{92.53}&&&&&\textbf{77.15}&\textbf{79.67}&\textbf{91.73}&\textbf{91.64}&\textbf{91.22}&\textbf{88.90}&
				\textbf{79.07}&\textbf{78.12}&\textbf{75.82}&\textbf{74.82}&\textbf{82.93}&\textbf{82.02}&\textbf{78.98}&\textbf{80.81}\\
				\midrule\multirow{3}{*}{WildReceipts~\cite{wildreceipts-sun2021}}
				& \robertabase{} &89.19&\multirow{3}{*}{$\approx7.03$}&\multirow{3}{*}{$\approx18.93$}&\multirow{3}{*}{$\approx7.17$}&\multirow{3}{*}{$\approx17.97$}&79.01&81.37&88.32&87.98&87.48&86.18&78.76&78.62&78.67&78.93&81.38&81.21&80.58&81.12\\
				& \layoutlmbase{} (w/o Image) &92.34&&&&&79.28&84.46&91.85&91.74&91.31&89.53&82.24&81.82&82.07&\textbf{85.46}&85.14&85.81&85.72&85.46\\
				& \layoutlmbase{} &\textbf{92.99}&&&&&\textbf{80.97}&\textbf{85.85}&\textbf{92.23}&\textbf{91.89}&\textbf{91.71}&\textbf{90.69}&\textbf{83.79}&\textbf{83.71}&\textbf{82.70}&83.31&\textbf{86.65}&\textbf{87.00}&\textbf{86.62}&\textbf{86.42}
				\\
				\midrule\multirow{2}{*}{XFUND~\cite{xfund}}
				& \layoutlmbasec{} (w/o Image) &\textbf{89.74}&\multirow{2}{*}{$\approx6.95$}&\multirow{2}{*}{$\approx19.40$}&\multirow{2}{*}{$\approx7.05$}&\multirow{2}{*}{$\approx18.07$}&\textbf{69.51}&\textbf{75.01}&86.59&\textbf{87.35}&\textbf{84.32}&\textbf{80.01}&\textbf{69.74}&\textbf{70.26}&\textbf{71.11}&\textbf{69.46}&\textbf{76.65}&\textbf{75.35}&\textbf{74.60}&74.52\\
				& \layoutlmbasec{} &89.43&&&&&69.23&74.93&\textbf{87.08}&86.08&82.65&77.07&68.64&68.50&63.40&67.59&75.18&75.00&71.93&\textbf{75.22}\\
				\midrule\multirow{3}{*}{DOCILE~\cite{docile}*}
				& \robertabase{} &75.13&\multirow{3}{*}{$\approx7.15$}&\multirow{3}{*}{$\approx18.89$}&\multirow{3}{*}{$\approx7.28$}&\multirow{3}{*}{$\approx18.28$}&53.19&59.39&-&-&71.30&67.76&-&-&50.80&53.83&-&-&58.92&59.39\\
				& \layoutlmbasec (w/o Image)&\textbf{81.59}&&&&&69.29&70.58&-&-&79.39&\textbf{78.55}&-&-&68.63&68.39&-&-&71.30&71.04\\
				& \layoutlmbasec &80.29&&&&&\textbf{69.89}&\textbf{73.19}&-&-&\textbf{79.75}&78.22&-&-&\textbf{70.45}&\textbf{69.97}&-&-&\textbf{72.76}&\textbf{72.95}\\
				\bottomrule
			\end{tabular}
		}\\
	\raggedright \tiny \qquad *Reported results were computed on the validation set due to unavailability of test set.
	\end{center}
\vspace{-1em}
	\caption{Performance comparison of different privacy methods on target datasets. We only report the results obtained at the end of the full training rather than reporting the best performance obtained over all training rounds/epochs. For DP experiments presented here, max sequence length was set to $L=128$, TSP was applied wherever possible, and the hyperparameters were chosen as described in \cref{sec:dp-hyper-param-tuning}. Since DOCILE~\cite{docile} is a much larger dataset, we did not feel it necessary to experiment with a small number of clients for this dataset and therefore we computed results only with $K{\in}\{8,16\}$. See \cref{app:final_training_params} for full configuration of hyperparameters used in these experiments.}
	\label{table:perf-eval}
	\vspace{-1em}
\end{table*}
\subsubsection{Assessing Parameter Reduction Techniques}
Training the models with DP-SGD/Adam is considerably compute and memory intensive, and therefore, we also investigated different parameter reduction methods for fine-tuning under DP. In particular, we explored training the models with LoRA~\cite{dp-lora-Yu2021}, fine-tuning only the top 4 (top4) layers of the model and fine-tuning only the top 8 (top8) layers of the model each of which trained only 1\%, 20\% and 50\% of the model parameters, respectively. We compare these methods against full fine-tuning (full) for two datasets FUNSD and CORD in \cref{fig:param-red}. In general, it was observed that full fine-tuning performed best when compared with other fine-tuning approaches, while top8 produced performances similar to full fine-tuning with only 50\% of the parameters trained. 
LoRA, on the other hand, generally did not perform well on our small-scale datasets even though it has previously been shown to perform well for NLP tasks on large datasets \cite{dp-Yu2021a,dp-Li2021}. 

\begin{guidelines}
Based on above observations, our general recommendation is that when sufficient compute resources are available, full finetuning should be preferred in order to achieve the best model performance.
\end{guidelines}

\subsection{Performance Evaluation - DP/FL/DP-FL}
In this section, we present the results of non-private fine-tuning, private fine-tuning via DP-Adam, private fine-tuning via FL, and the proposed private fine-tuning via FeAm-DP applied to the LayoutLMv3 model. To train the models with DP-Adam, the appropriate parameters were selected according to the guidelines provided in \cref{sec:dp-hyper-param-tuning}, and the exact same parameters were used for DP-FL. For FL, no additional hyperparameter tuning was performed. The results are summarized in \cref{table:perf-eval} in which we present the performance (F1-score) of the \layoutlmbase{} model under two settings, \layoutlmbase{} with image, and \layoutlmbase{} with without image (which uses only text and layout information) and compare it with the text-only baseline model \robertabase{} on the six datasets described in \cref{sec:datasets}. We train both \layoutlmbase{} (w/o image) and RoBERTa under the same hyperparameter settings as \layoutlmbase{} for all scenarios.  
In addition to performance, for each dataset, we also report the converted $\epsilon$ values under Gaussian Accountant $\epsilon_\text{G}$ and PRV accountant $\epsilon_\text{PRV}$ both of which result in slightly better privacy (lower $\epsilon$) than the RDP accountant $\epsilon$.

A number of important observations can be made from the results reported in \cref{table:perf-eval}. For most datasets, even after considerable hyperparameter tuning, DP resulted in a significant loss of model utility in exchange for increased privacy. 
Even so, we were still able to achieve reasonable performance for the KIE task, ranging from $69\%$ to $80\%$ for privacy budget $\epsilon=8$, and from $73\%$ to $86\%$ for privacy budget $\epsilon=20$. 
In addition, the performance gap between non-private baseline models and private models decreased with increasing dataset size (as evident by results on WildReceipts and DOCILE datasets). Another noticeable aspect is that on some datasets such as FUNSD and SROIE, the additional image modality information did not present huge performance improvements in non-private training. However, in the case of private training ($\epsilon\in\{8,20\}$), it resulted in some significant improvements. Contrary to this, adding additional image information to XFUND adversely affected performance. Furthermore, LayoutLMv3 significantly outperformed the text-only baseline RoBERTa on all datasets under private training, despite the performance gap between the two being much smaller under non-private training. 

In contrast to DP, FL under different sets of clients $K\in\{2,4,8,16\}$, resulted in only minor loss of performance $(2\%{\sim}8\%)$ which was increased as the number of clients were increased. This was expected as greater number of clients provide more privacy due to sampling only a fraction $C$ of clients in each FL round. Finally, in a FL-DP setting, using the proposed FeAm-DP algorithm, we observed that overall the performance of the models were similar to standalone DP as expected and even increasing the number of clients had only a minor effect on the performance loss. Interestingly, in some cases such as on WildReceipts and CORD datasets, FeAm-DP resulted in an even higher performance than the standalone DP.

\subsection{Conclusion}
In this paper, we proposed strategies for developing private document KIE systems by leveraging large pretrained document foundation models in combination with 3 key privacy methods: differential privacy (DP) and federated learning (FL), and differentially private federated learning (DP-FL). Through an extensive evaluation over six benchmark datasets, we demonstrated that large document foundation models can be effectively fine-tuned for the KIE task with sufficient utility under strong privacy guarantees ($\epsilon\in\{8,20\}$ for DP and DP-FL). In addition, based on a comprehensive analysis of the effects of various training parameters on private fine-tuning, we presented practical guidelines for achieving an optimal privacy-utility tradeoff under global DP. Lastly, we proposed FeAm-DP, an algorithm for efficient upscaling of standalone DP to multiclient federated setting and demonstrated its effectiveness through a range of experiments. Overall, our work mainly focused on the application of global DP for the KIE task. In future, it will also be worthwhile to explore local DP-based approaches for private KIE.

{\small
	\bibliographystyle{ieee_fullname}
	\bibliography{references}
}

\clearpage
\clearpage
\appendix

\section{DP-SGD/Adam}
In our experiments, we use DP-Adam to fine-tune our models under global DP. DP-Adam functions exactly like DP-SGD, except that it uses Adam optimizer instead of SGD optimizer. For completeness, We have included the pseudocode for both algorithms in \cref{alg:dpsgd}.
\SetKwProg{ClientUpdate}{ClientUpdate}{\textbf{:}}{\KwRet{$\mathbf{\tilde{g}}_{k,t}$}}
\begin{algorithm}[h]\footnotesize\SetAlgoLined
\caption{DP-SGD/Adam}
\KwIn{$\mathcal{L}(\theta) = \frac{1}{B}\sum_i\mathcal{L}(\theta, x_i)$, Dataset $\mathcal{D}={(x_1,y_1),\dots,(x_N,y_N)}$, learning rate $\eta$, gradient clipping norm $S$, noise scale $\sigma$, sampling rate $q$, target $(\epsilon,\delta)$, privacy accountant $\mathcal{M}$, total training steps $T$}
\KwInit{Initialize $\theta_0$ randomly}
\For{each step $t=1,\dots,T$}{
$\mathcal{B} \gets ($sample a batch of size $B$ with sampling probability $q$)\\
\ForEachGradient{$x_i\in\mathcal{B}$}{
	// Compute gradient\\
	$\mathbf{g}(x_i) \gets \nabla_{\theta_t} \mathcal{L}(\theta_t, x_i)$\\
	// Clip gradient\\
	$\mathbf{\bar{g}}(x_i) \gets \mathbf{g}(x_i) / max(1, \frac{||\mathbf{\bar{g}}(x_i)||_2}{S})$}
// Add noise\\
$\mathbf{\tilde{g}} \gets \frac{1}{B}(\sum_i\mathbf{\bar{g}}(x_i) + \mathcal{N}(0, \sigma^2S^2\mathbf{I})$\\
\uIf{Algorithm is DP-SGD}{
	// Call SGD Update \\
	$\theta_{t+1} \gets \theta_t - \eta \mathbf{\tilde{g}}$
}
\ElseIf{Algorithm is DP-Adam}{
	// Call Adam Update \\
	$m_t \gets \beta_1m_{t-1} + (1-\beta_1)\mathbf{\tilde{g}}$\\
	$v_t \gets \beta_2v_{t-1} + (1-\beta_2)\mathbf{\tilde{g}}^2$\\
	$m_t \gets \frac{m_{t}}{1-\beta_1^t}$\\
	$v_t \gets \frac{v_{t}}{1-\beta_2^t}$\\
	$\theta_{t+1} \gets \theta_t + \eta\frac{m_t}{\sqrt{v_t}+\tau} $
}
print  $\mathcal{M}.\text{get\_privacy\_spent}(q,\sigma,t,\delta)$
}
\label{alg:dpsgd}
\end{algorithm}
\label{app:dp-adam}
\section{FedAvg Algorithm}
For training the models in federated learning settings, we utilize the FedAvg algorithm as described in \cref{exp:fl}. The pseudocode for this algorithm is available in \cref{alg:fedavg}. For all FL experiments, the number of local epochs per client is set to one ($E=1$) for each communication round and the models are trained using the Adam optimizer.
\SetKwProg{ClientUpdate}{ClientUpdate}{\textbf{:}}{\KwRet{$\theta$}}
\begin{algorithm}[!t]\footnotesize\SetAlgoLined
	\caption{FedAvg}
	\KwIn{Learning rate $\eta$, total clients K, clients sampling rate $C$, total FL rounds $T$}
\Server{
	\KwInit{$\theta_0$, $m \gets CK$}
	\For{each round $t=1,\dots,T$}{
		$\mathcal{S}_t \gets (\text{sample a set of $m$ clients from $K$})$\\
		\ForEachParallel{each client $k\in\mathcal{S}$}{
			$\mathbf{\theta}_{k,t} \gets \text{ClientUpdate}(k, \theta_{t-1})$
		}
		$\theta_{t} \gets \sum_{k\in\mathcal{S}_t}\frac{n_k}{n}\theta_{k,t}$\\	
	}
}	
\ClientUpdate{($k, \theta$)}{
	\KwIn{$\mathcal{L}(\theta) = \frac{1}{B}\sum_i\mathcal{L}(\theta, x_i)$, $\mathcal{D}_k$ of size $|\mathcal{D}_k|$}
	$\mathcal{B}\gets ($sample a batch of size $B$)\\
	\For{each epoch $e=1,\dots,E$}{
		\ForEachGradient{$b\in\mathcal{B}$}{
				\uIf{Optimizer is SGD}{
					// Call SGD Update \\
					$\theta \gets \theta - \eta \nabla_{\theta} \mathcal{L}(\theta, b)$\\
				}
				\ElseIf{Optimizer is Adam}{
					// Call Adam Update \\
					$\mathbf{\tilde{g}} \gets\nabla_{\theta} \mathcal{L}(\theta, b)$\\
					$m_t \gets \beta_1m_{t-1} + (1-\beta_1)\mathbf{\tilde{g}}$\\
					$v_t \gets \beta_2v_{t-1} + (1-\beta_2)\mathbf{\tilde{g}}^2$\\
					$m_t \gets \frac{m_{t}}{1-\beta_1^t}$\\
					$v_t \gets \frac{v_{t}}{1-\beta_2^t}$\\
					$\theta \gets \theta - \eta \frac{m_t}{\sqrt{v_t}+\tau}$\\
				}
		}
	}
}
\label{alg:fedavg}
\end{algorithm}
\label{app:fedavg}
\section{LayoutLMv3 for KIE}
\begin{figure}[!t]
	\centering	
	\includegraphics[width=\linewidth]{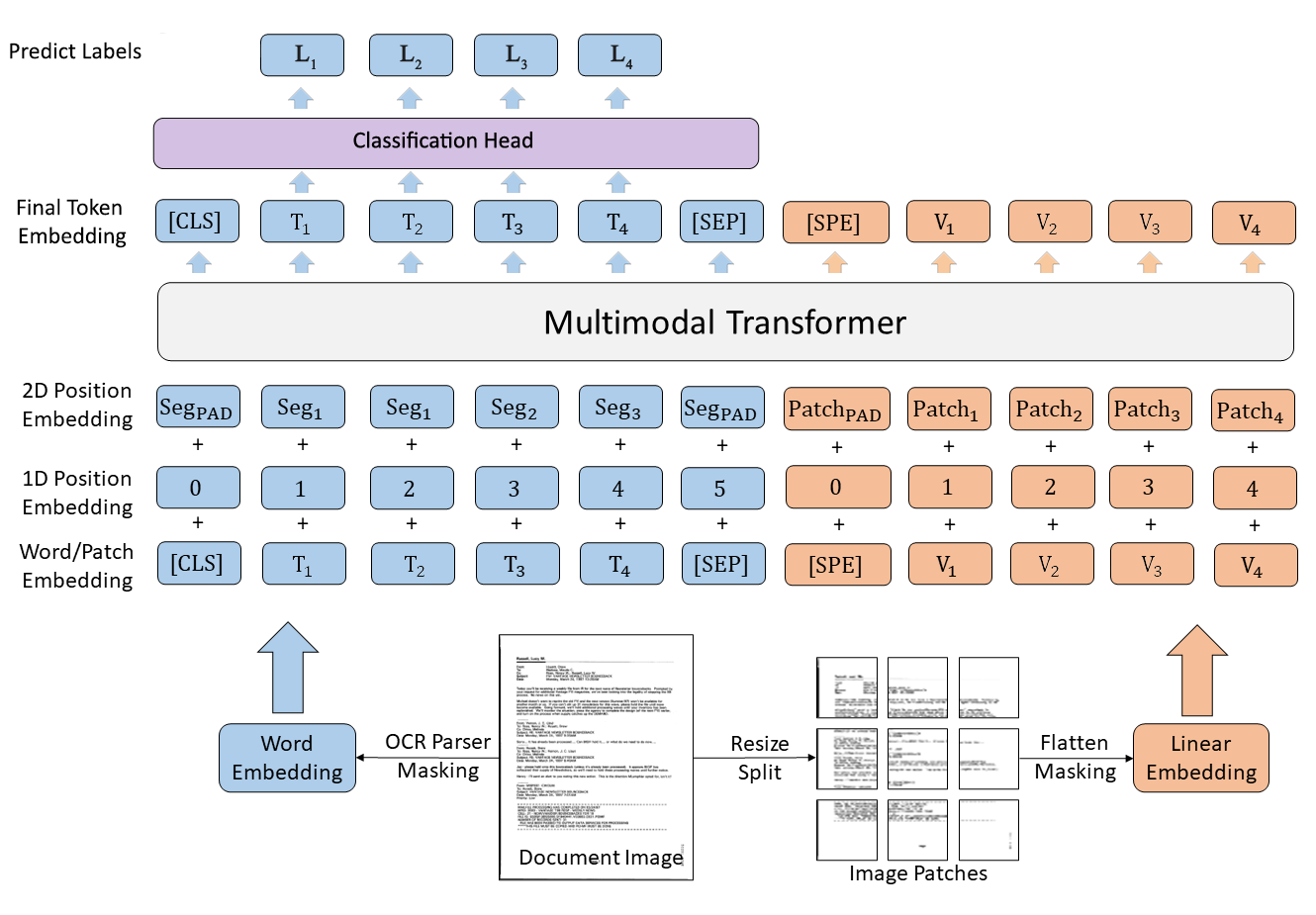}
	\caption{LayoutLMv3 model architecture with token-level classification for key information extraction (KIE).}
	\label{fig:layoutlmv3}
\end{figure}

\cref{fig:layoutlmv3} illustrates the layoutlmv3 model architecture with a token-level classification head that we have used in this work. As shown, the model takes as input a tokenized text sequence and calculates the embedding of each token by adding together the text embedding with the 2D bounding box position embeddings, and the 1D position embeddings.  In the next step, the input text sequence is concatenated with an image embedding sequence, which is generated by converting the input image into fixed-sized patches. The input is then fed into a multimodal transformer network in order to generate the final embeddings at the token level. In the final step, a classification head is used to predict the label associated with each token embedding. It is important to note that LayoutLMv3 generates 2D position embeddings using segment-level layout (which we have also used in this study). As a result, the bounding box for each text token corresponds to the union of all the bounding boxes within a single segmented region such as a paragraph, table, etc. The result is that multiple words in the same segment have the same 2D layout embedding. 
\label{app:layoutlm}
\section{Full Model Configuration}
A complete list of the model configuration parameters used in our study can be found in \cref{tab:model-config}.
\begin{table}[b]
	\centering
	\begin{center}
		\resizebox{0.7\linewidth}{!}{
			\begin{tabular}{@{}ll@{}}
				\toprule
				Model Parameter&Configuration\\
				\midrule
				Number of encoder layers&$12$\\
				Number of self-attention heads&$12$\\
				Embedding dimension & $768$\\
				Image input resolution $C\times H\times W$& $3\times224\times224$\\
				Image normalization mean&$[0.5, 0.5, 0.5]$\\
				Image normalization std&$[0.5, 0.5, 0.5]$\\
				Image patch size $P\times P$& $16\times 16$\\
				Dropout & $\{0,0.1\}$\\
				Text sequence Length & $\{128,512\}$\\
				Image sequence Length & $198$\\
				Text lower-cased & True\\
				Tokenization stride & $0$\\
				Segment-level layout & True\\
				\bottomrule
			\end{tabular}
		}
	\end{center}
	\caption{The full model configuration parameters used in our study.}
	\label{tab:model-config}
\end{table}

\label{app:model-config}
\section{Privacy Accounting}
\label{app:privacy_accounting}
To account for privacy loss $(\epsilon)$, we use the following accountants in our work: (1) R\'enyi DP~\cite{dp-rdp-mironov}, (2) Gaussian DP~\cite{dp-gdp-koskela2022individual}, and (3) Private Random Variable (PRV) Accountant~\cite{dp-prv-gopi}, all of which represent an improvement over the moments accountant proposed by Abadi \etal~\cite{dpsgd-Abadi2016}. 
Accounting for privacy loss with R\'enyi DP provides a strict upper bound however it may overestimate the privacy loss. On the other hand, the Gaussian Accountant only provides an approximation to the actual loss, which may also lead to an underestimation of the loss. To counter these problems, Gopi \etal~\cite{dp-prv-gopi} proposed the PRV Accountant which gives a better numerical approximation of the privacy leakage compared to  R\'enyi DP and Gaussian DP accountants.
 
Given a sampling rate $q$, and the target privacy budget $(\epsilon,\delta)$, privacy accountants can be used to estimate the privacy leakage over a fixed number of epochs $E$ for a given noise scale $\sigma$ and this process can be numerically optimized to obtain a suitable $\sigma$ for a target privacy budget $(\epsilon,\delta)$. We perform this optimization in combination with the R\'enyi DP~\cite{dp-rdp-mironov} throughout our experiments to compute the noise scale $\sigma$. Once the noise scale $\sigma$ is determined, we also use it to determine the privacy leakage estimate according to the Gaussian Accountant and the PRV Accountant for each dataset as reported in \cref{table:perf-eval}.

\begin{table}[!ht]
	\centering
	\begin{center}
		\resizebox{0.8\linewidth}{!}{
			\begin{tabular}{@{}lll@{}}
				\toprule
				Dataset&Field Type &Description\\
				\midrule
				\multirow{3}{*}{\parbox{2cm}{FUNSD/\\XFUND}}&header&Document Header\\
				&question&Text sequence representing a question\\
				&answer&Text sequence representing a answer\\
				&other&Others\\
				\midrule
				\multirow{30}{*}{CORD}&menu.nm&	Name of menu\\
				&menu.num&	Identification number of menu\\
				&menu.unitprice&	Unit price of menu\\
				&menu.cnt	&Quantity of menu\\
				&menu.discountprice&	Discounted price of menu\\
				&menu.price&	Total price of menu\\
				&menu.itemsubtotal&	Price of each menu after discount applied\\
				&menu.vatyn	&Whether the price includes tax or not\\
				&menu.etc&	Others\\
				&menu.sub\_nm&	Name of submenu\\
				&menu.sub\_unitprice	&Unit price of submenu\\
				&menu.sub\_cnt	&Quantity of submenu\\
				&menu.sub\_price	&Total price of submenu\\
				&menu.sub\_etc	&Others\\
				&void\_menu.nm	&Name of menu\\
				&void\_menu.price &Total price of menu\\
				&subtotal.subtotal\_price	&Subtotal price\\
				&subtotal.discount\_price	&Discounted price in total\\
				&subtotal.service\_price&	Service charge\\
				&subtotal.othersvc\_price&	Added charge other than service charge\\
				&subtotal.tax\_price&	Tax amount\\
				&subtotal.etc&Others\\
				&total.total\_price	&Total price\\
				&total.total\_etc	&Others\\
				&total.cashprice&	Amount of price paid in cash\\
				&total.changeprice&	Amount of change in cash\\
				&total.creditcardprice	&Amount of price paid in credit-debit card\\
				&total.emoneyprice	&Amount of price paid in emoney, point\\
				&total.menutype\_cnt	&Total count of type of menu\\
				&total.menuqty\_cnt	&Total count of quantity\\
				&other&Others\\
				\midrule
				\multirow{4}{*}{SROIE}&company&Name of the company\\
				&date&Date\\
				&address&Company address\\
				&total&Total receipt amount\\
				&other&Others\\
				\midrule
				\multirow{25}{*}{WildReceipts}
				&store\_name\_value&Store name\\
				&store\_name\_key&Store name key\\
				&store\_addr\_value&Store address\\
				&store\_addr\_key&Store address key\\
				&tel\_value&Phone number\\
				&tel\_key&Phone number key\\
				&date\_value&Date\\
				&date\_key&Date Key\\
				&time\_value&Time\\
				&time\_key&Time key \\
				&prod\_item\_value&Product item value\\
				&prod\_item\_key&Product item value key\\
				&prod\_quantity\_value&Product item quantity\\
				&prod\_quantity\_key&Product item quantity key\\
				&prod\_price\_value&Product price\\
				&prod\_price\_key&Product price key\\
				&subtotal\_value&Subtotal value\\
				&subtotal\_key&Subtotal key\\
				&tax\_value&Tax value\\
				&tax\_key&Tax key\\
				&tips\_value&Tips value\\
				&tips\_key&Tips key\\
				&total\_value&Total value\\
				&total\_key&Total key\\
				&other&Others\\
				\midrule
				\multirow{35}{*}{DOCILE}
				&account num&Bank account number\\
				&amount due &Total amount to be payed\\
				&amount paid &Total amount already paid\\
				&amount total gross &Total amount with tax\\
				&amount total net &Total amount without tax\\
				&amount total tax &Total sum of tax amounts\\
				&bank num &Bank number\\
				&bic &Bank Identifier Code (SWIFT)\\
				&currency code amount due &Currency code or symbol found near the amount due\\
				&customer billing address &Address of the company that is being invoiced\\
				&customer billing name &Name of the company that is being invoiced\\
				&customer delivery address &Address of the company for delivery of goods/services\\
				&customer delivery name &Name of the company for delivery of goods/services\\
				&customer id &Customer account number\\
				&customer order id &Any customer order reference\\
				&customer other address &Any other name and address of the purchasing company\\
				&customer other name &Any other name of the purchasing company\\
				&customer registration id &Purchaser registration identifier number\\
				&customer tax id &Purchaser tax identification number\\
				&date due &Due date for payment\\
				&date issue &Date the document was issued\\
				&document id &Main document number\\
				&iban &International Bank Account Number\\
				&order id &Any order number\\
				&payment reference &Payment reference number\\
				&payment terms &Conditions for the payment time window\\
				&tax detail gross &Tak breakdown line amount with tax\\
				&tax detail net &Tax breakdown line amount without tax\\
				&tax detail rate &Tax breakdown line tax rate\\
				&tax detail tax &Tax breakdown line tax amount\\
				&vendor address &Address of the supplier company\\
				&vendor email &Any supplier e-mail address\\
				&vendor name& Name of the supplier company\\
				&vendor order id &Any vendor order reference\\
				&vendor registration id &Supplier registration identification number\\
				&vendor tax id &Supplier tax identification number\\
				&other&Others\\
				\bottomrule
			\end{tabular}
		}
	\end{center}
	\caption{Full list of target entity types for the datasets investigated in our study.}
	\label{tab:entities}
	\vspace{-1em}
\end{table}
\begin{figure}[ht]
\centering	
\begin{subfigure}{\linewidth}
	\includegraphics[width=\linewidth]{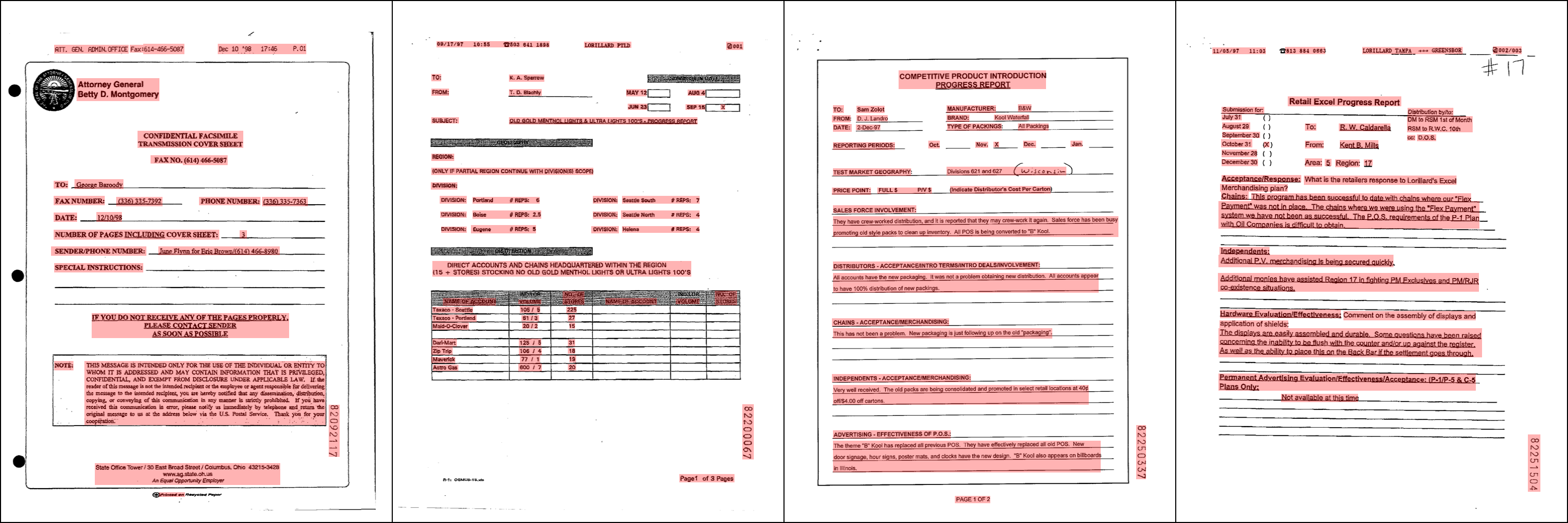}
	\caption{FUNSD}
\end{subfigure}
\begin{subfigure}{\linewidth}
	\includegraphics[width=\linewidth]{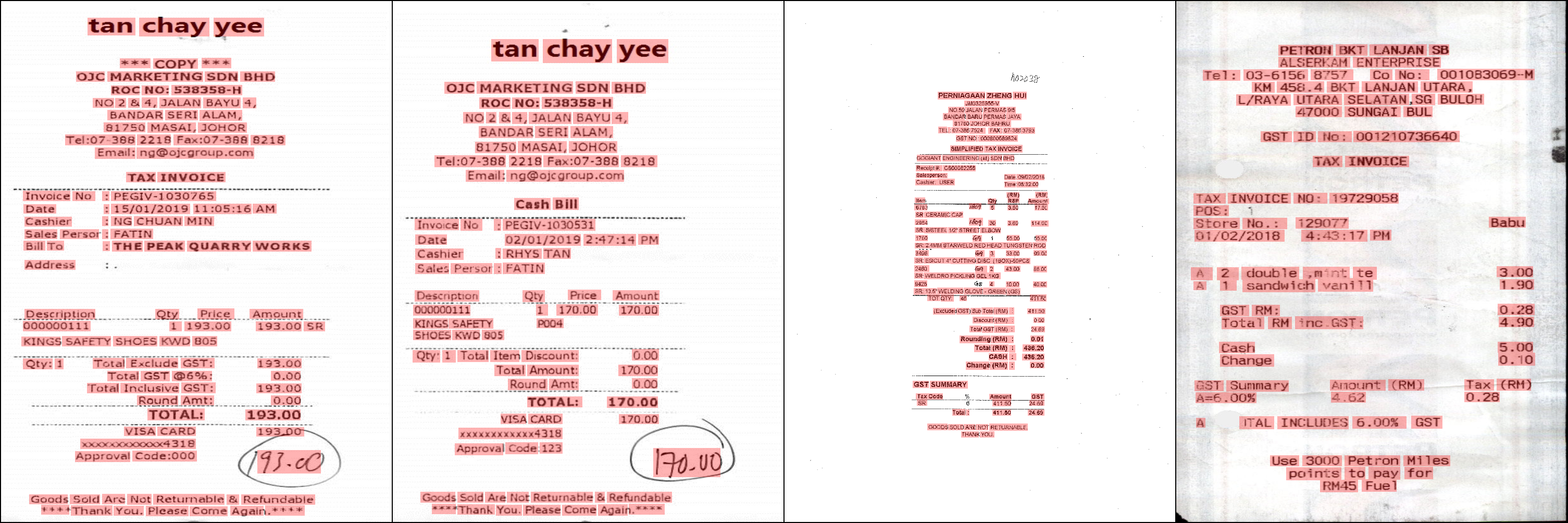}
	\caption{SROIE}
\end{subfigure}
\begin{subfigure}{\linewidth}
	\includegraphics[width=\linewidth]{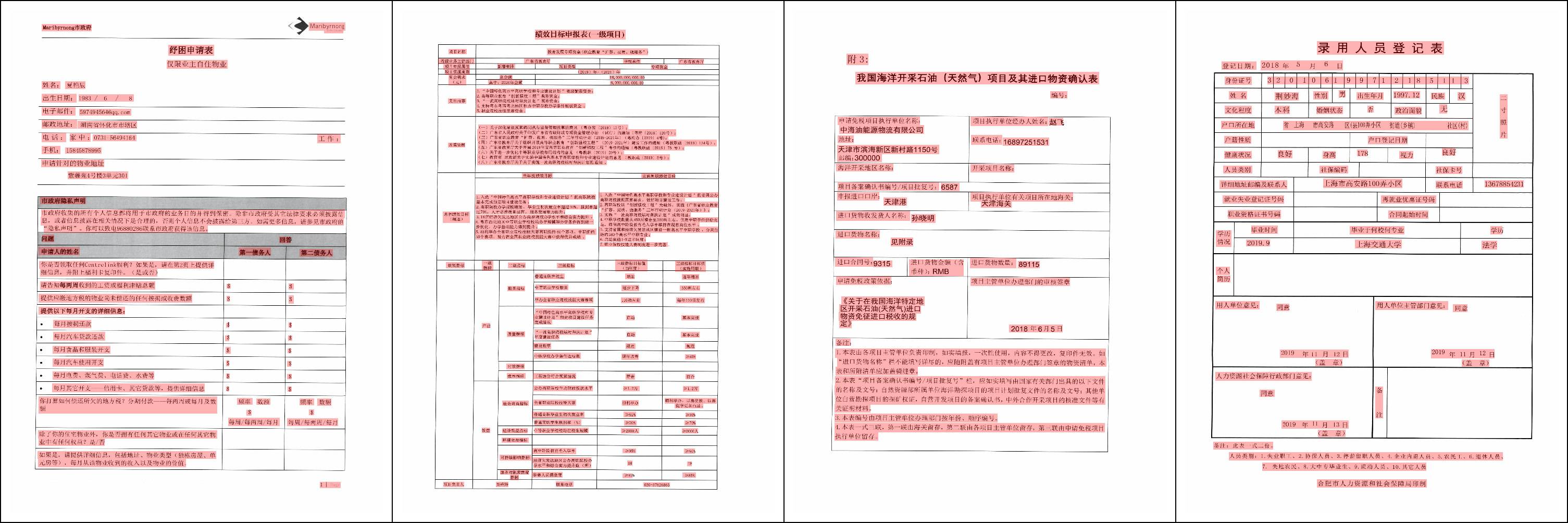}
	\caption{XFUND}
\end{subfigure}
\begin{subfigure}{\linewidth}
	\includegraphics[width=\linewidth]{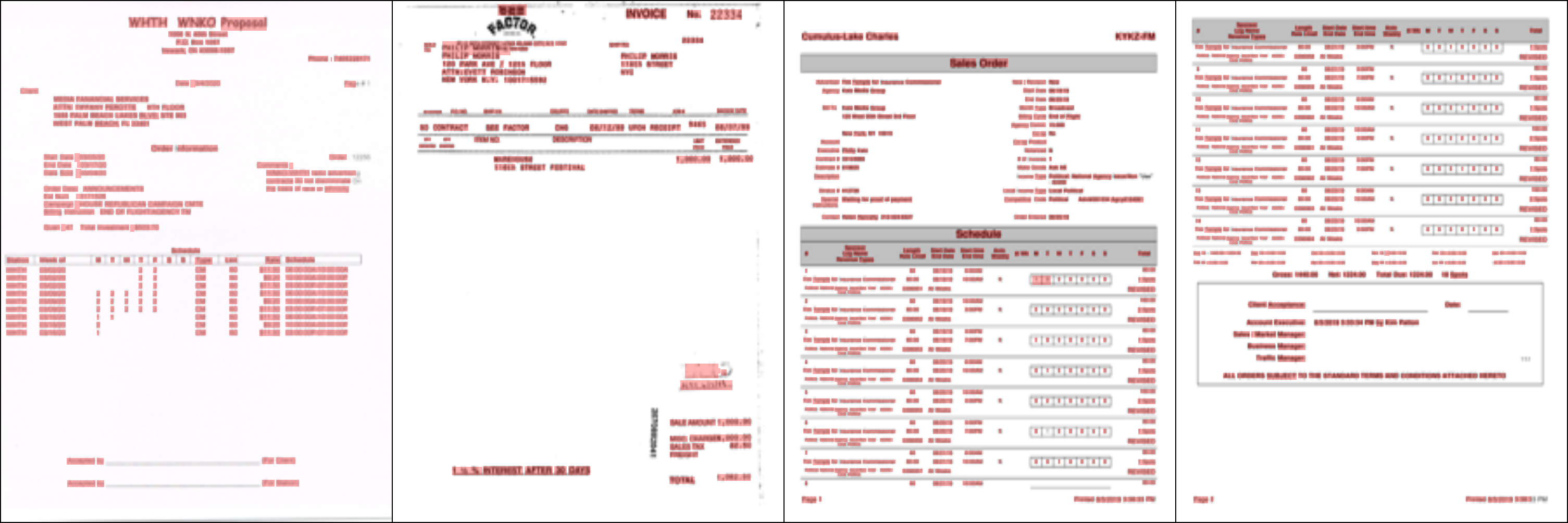}
	\caption{DOCILE}
\end{subfigure}
\caption{A few samples from the different datasets are shown.}
\label{fig:dataset_examples}
\end{figure}

\begin{table}[!ht]
	\centering
	\begin{center}
		\resizebox{\linewidth}{!}{
			\begin{tabular}{@{}ll@{}}
				\toprule
				Parameter& Value\\
				\midrule
				DP guarantee $(\epsilon,\delta)$&$(20,1/|\mathcal{D}_{train}|)$\\
				Clipping norm (S) & $0.1$\\
				Learning rate $(\eta)$& $1.0e-3$\\
				Learning rate decay & False\\
				Epochs $(E)$ & $40$ for non-private;$50$ for private\\
				Weight decay $(\lambda)$ & 0\\
				\multirow{2}{*}{Noise scale $(\sigma)$} & \multirow{2}{*}{\parbox{5cm}{computed such that privacy budget $(\epsilon,\delta)$ is spent after $E$ epochs}}\\&\\
				Client sampling rate (C)&1.0 for $K=2$, 0.5 for $K>2$\\
				Total federated rounds $(T)$ & $40$ for FL;$\frac{1}{q}50$ for DP-FL\\
				Fine-tuning strategy&Full\\
				\bottomrule
			\end{tabular}
		}
	\end{center}
	\caption{Full list of default hyperparameters that were used in our study.}
	\label{tab:default-hyperparams}
\end{table}
\begin{table}[!hb]
	\centering
	\begin{center}
		\resizebox{0.7\linewidth}{!}{
			\begin{tabular}{@{}lll@{}}
				\toprule
				Dataset&Parameters& Values\\
				\midrule
				\multirow{7}{*}{FUNSD}
				&Sequence length $(L)$&$128$\\
				&Batch size $(B)$ & $64$\\
				&Sampling rate $(q)$ & $\frac{1}{7}$\\
				&Clipping norm (S) & $0.1$\\
				&Learning rate $(\eta)$& $5.0e-4$\\
				&Dropout&$0.1$\\
				&TSP Dataset & CORD\\
				&Fine-tuning strategy&Full\\
				\midrule
				\multirow{7}{*}{CORD}
				&Sequence length $(L)$&$128$\\
				&Batch size (B) & $384$\\
				&Sampling rate (q) & $\frac{1}{3}$\\
				&Clipping norm (S) & $0.1$\\
				&Learning rate $(\eta)$& $5.0e-4$\\
				&Dropout&$0.0$\\
				&TSP Dataset & WildReceipts\\
				&Fine-tuning strategy&Full\\
				\midrule
				\multirow{7}{*}{SROIE}
				&Sequence length $(L)$&$128$\\
				&Batch size (B) & $384$\\
				&Sampling rate (q) & $\frac{1}{5}$\\
				&Clipping norm (S) & $0.1$\\
				&Learning rate $(\eta)$& $5.0e-4$\\
				&Dropout&$0.0$\\
				&TSP Dataset & DOCILE-Synthetic\\
				&Fine-tuning strategy&Full\\
				\midrule
				\multirow{7}{*}{WildReceipts}
				&Sequence length $(L)$&$128$\\
				&Batch size $(B)$ & $384$\\
				&Sampling rate $(q)$ & $\frac{1}{6}$\\
				&Clipping norm $(S)$ & $0.1$\\
				&Learning rate $(\eta)$& $5.0e-4$\\
				&Dropout&$0.0$\\
				&TSP Dataset & CORD\\
				&Fine-tuning strategy&Full\\
				\midrule
				\multirow{7}{*}{XFUND}
				&Sequence length $(L)$&$128$\\
				&Batch size $(B)$ & $384$\\
				&Sampling rate $(q)$ & $\frac{1}{9}$\\
				&Clipping norm $(S)$ & $0.1$\\
				&Learning rate $(\eta)$& $5.0e-4$\\
				&Dropout&$0.1$\\
				&TSP Dataset & None\\
				&Fine-tuning strategy&Full\\
				\bottomrule
				\multirow{7}{*}{DOCILE}
				&Sequence length $(L)$&$512$\\
				&Batch size $(B)$ & $3256$\\
				&Sampling rate $(q)$ & $\frac{1}{3}$\\
				&Clipping norm $(S)$ & $0.1$\\
				&Learning rate $(\eta)$& $5.0e-4$\\
				&Dropout&$0.0$\\
				&TSP Dataset & DOCILE-Synthetic\\
				&Fine-tuning strategy&Full\\
				\bottomrule
			\end{tabular}
		}
	\end{center}
	\caption{Final training hyperparameters that were used to fine-tune the models under DP and DP-FL to get the the results reported in \cref{table:perf-eval}.}
	\label{tab:final-hyperparams}
\end{table}
\section{Dataset Details}
In \cref{fig:dataset_examples}, we provide a few samples from the datasets that were used in our study. As can be seen, the datasets show a lot of diversity in terms of sample types. In \cref{tab:entities}, we summarize the types of target entities available in each dataset. As can be seen from the comparison between CORD and WildReceipts, the two datasets have a great deal in common in terms of entity types (in addition to the fact that both datasets are receipts datasets). Among the similar entities are product names, quantities, and values, as well as totals, subtotals, and tax prices. 
\begin{figure*}[!t]
	\centering	
	\includegraphics[width=0.8\linewidth]{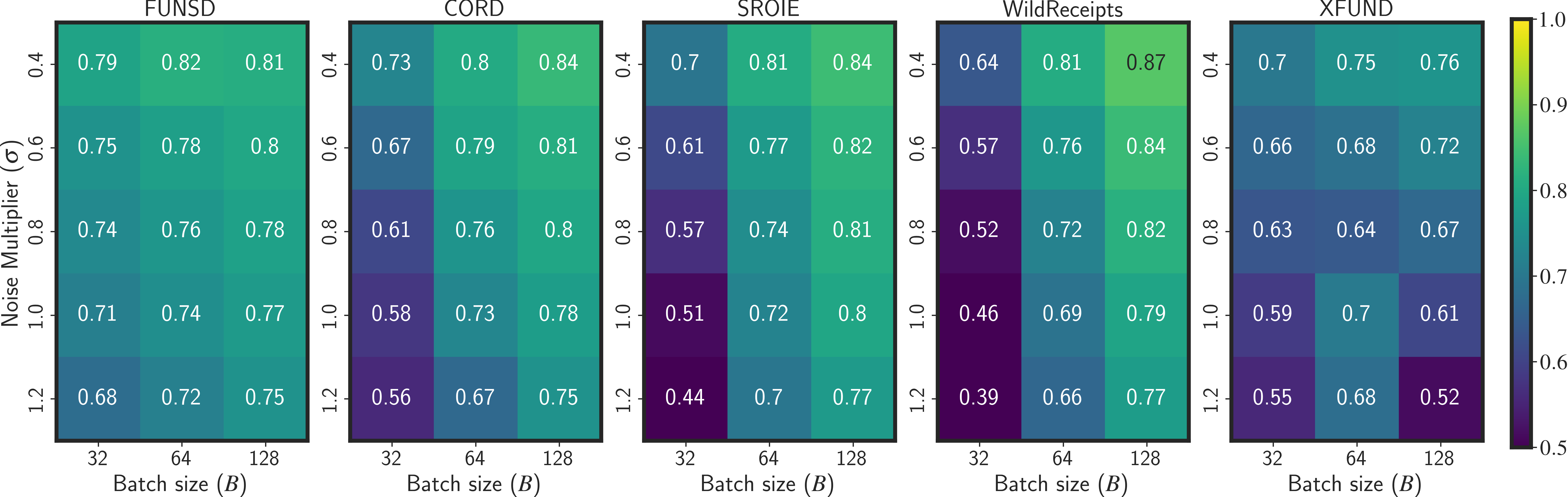}
	\caption{Full results showing the effect of varying batch size and noise multiplier on multiple datasets. The results were obtained with $\epsilon=20$ and $L=512$.}
	\label{fig:add_res_nm_vs_bs}
\end{figure*}
\begin{figure*}[!t]
\centering	
\begin{subfigure}{0.25\linewidth}
	\includegraphics[width=\linewidth]{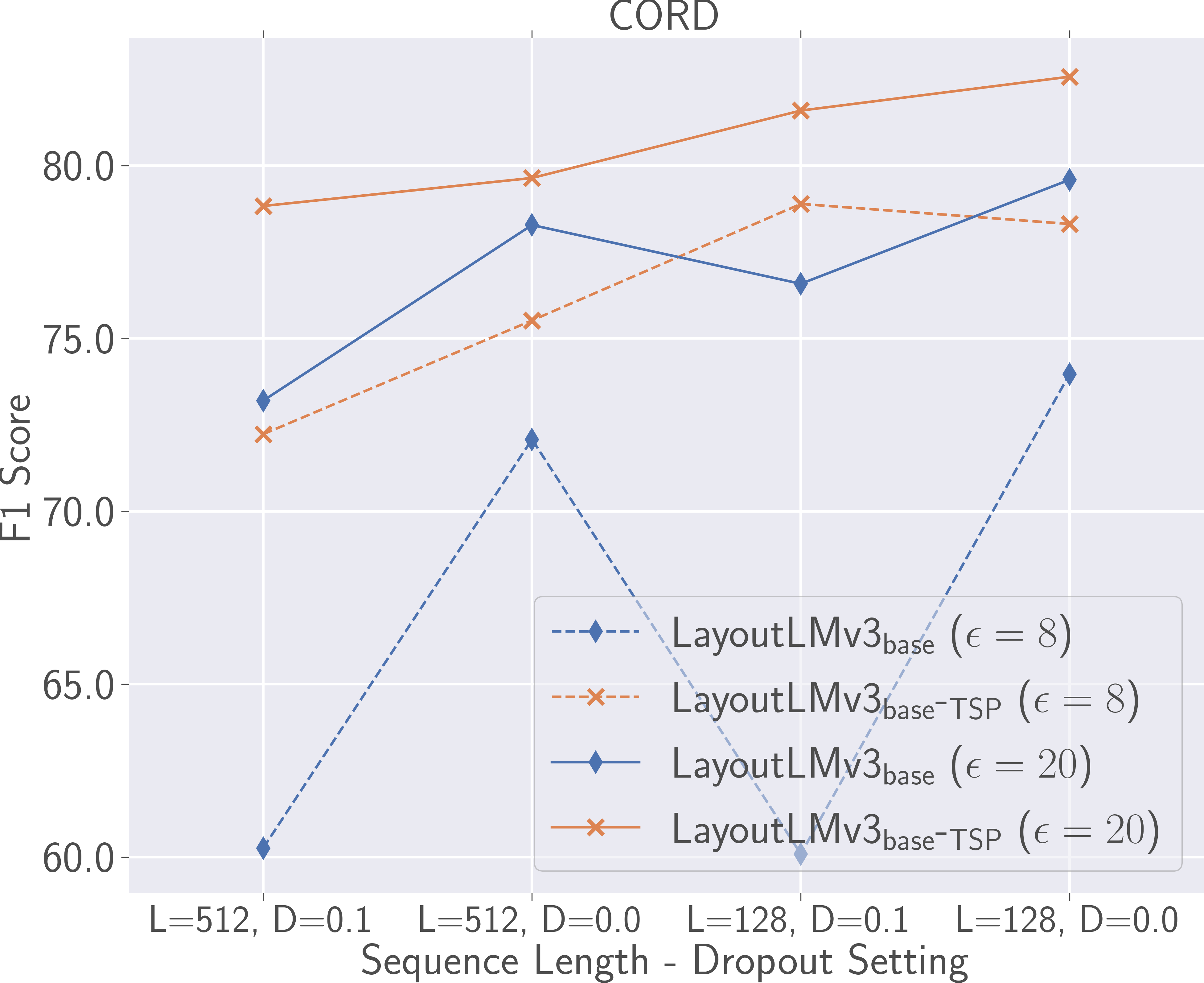}
\end{subfigure}
\begin{subfigure}{0.25\linewidth}
	\includegraphics[width=\linewidth]{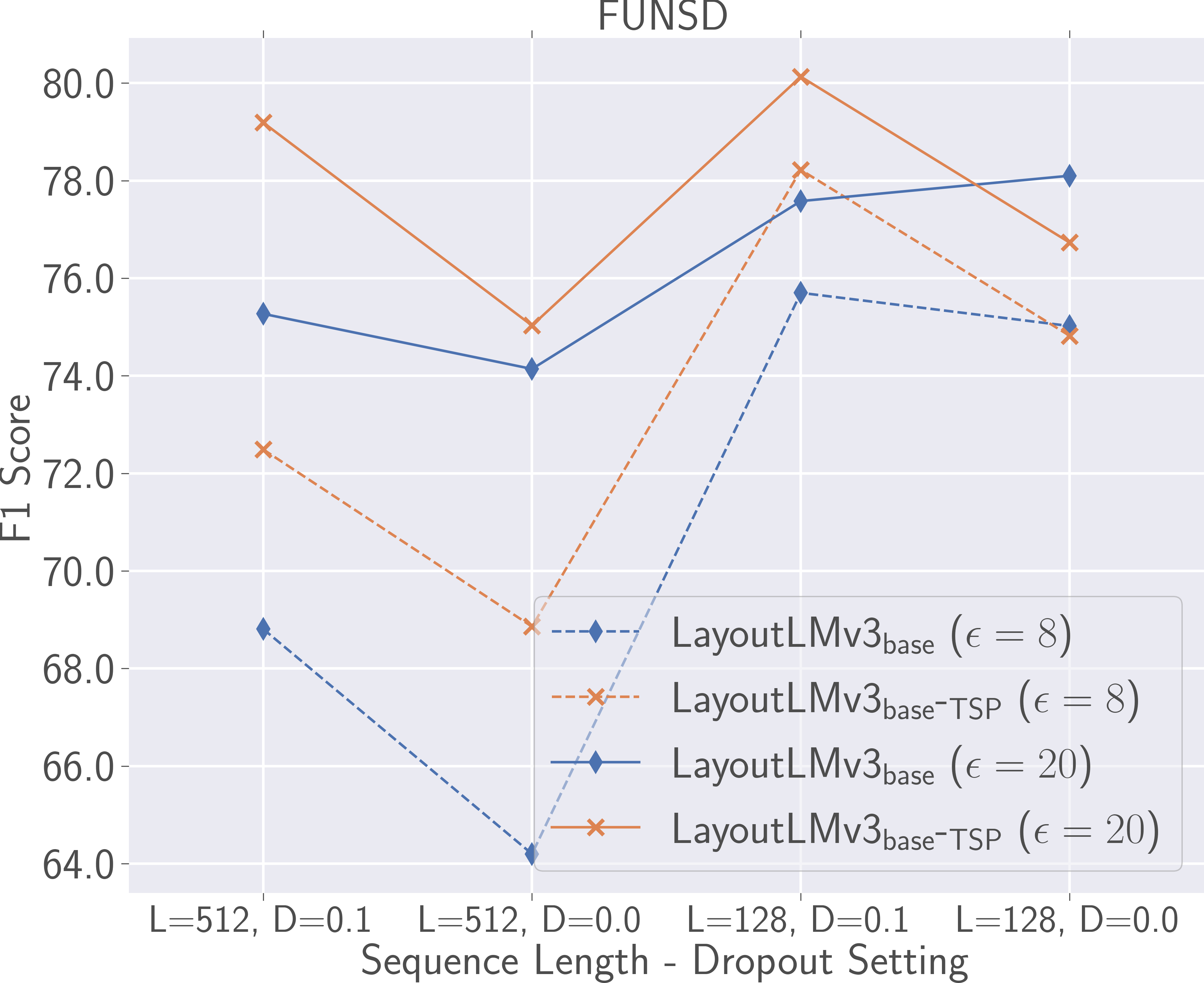}
\end{subfigure}
\begin{subfigure}{0.25\linewidth}
	\includegraphics[width=\linewidth]{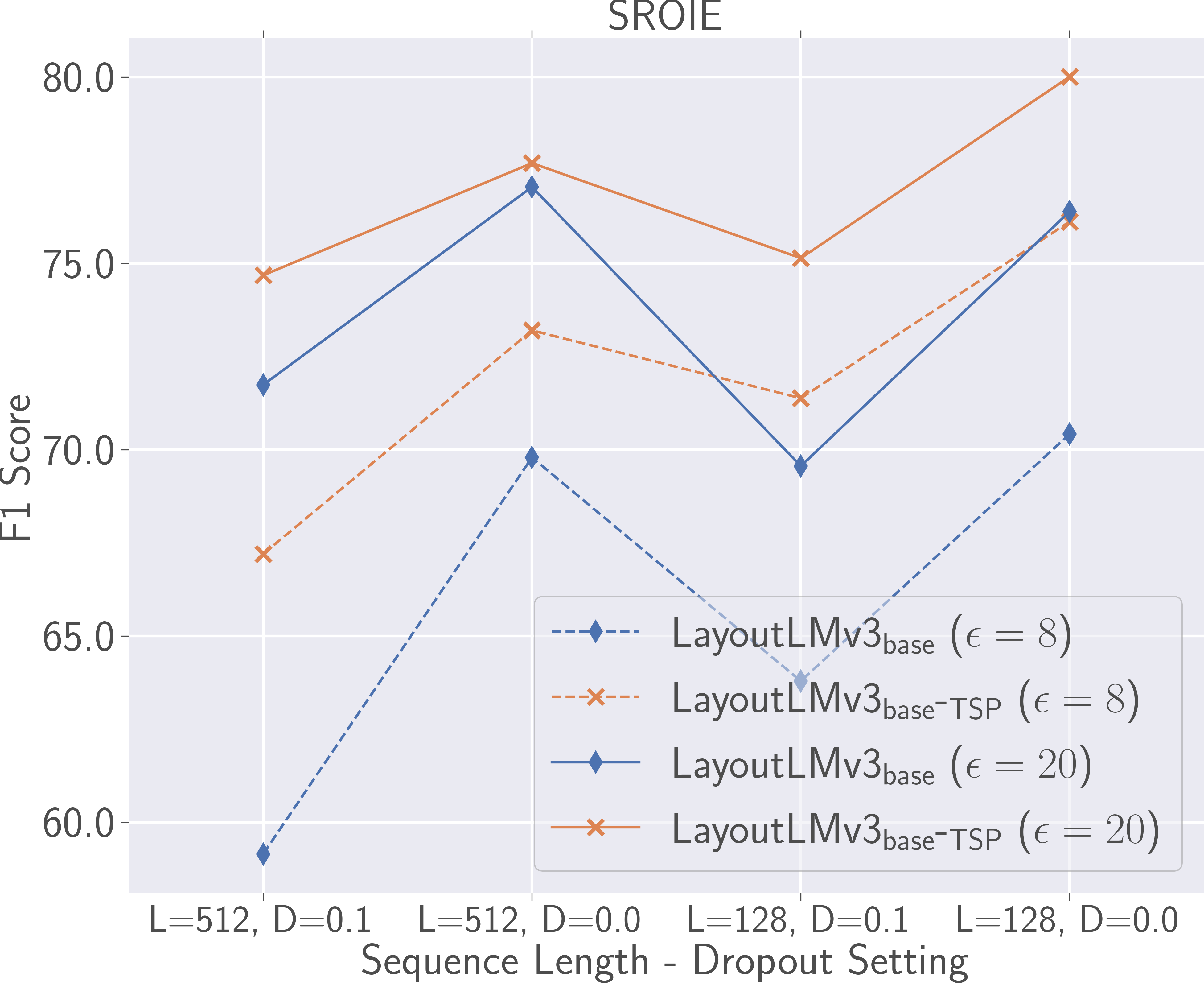}
\end{subfigure}\\
\begin{subfigure}{0.25\linewidth}
	\includegraphics[width=\linewidth]{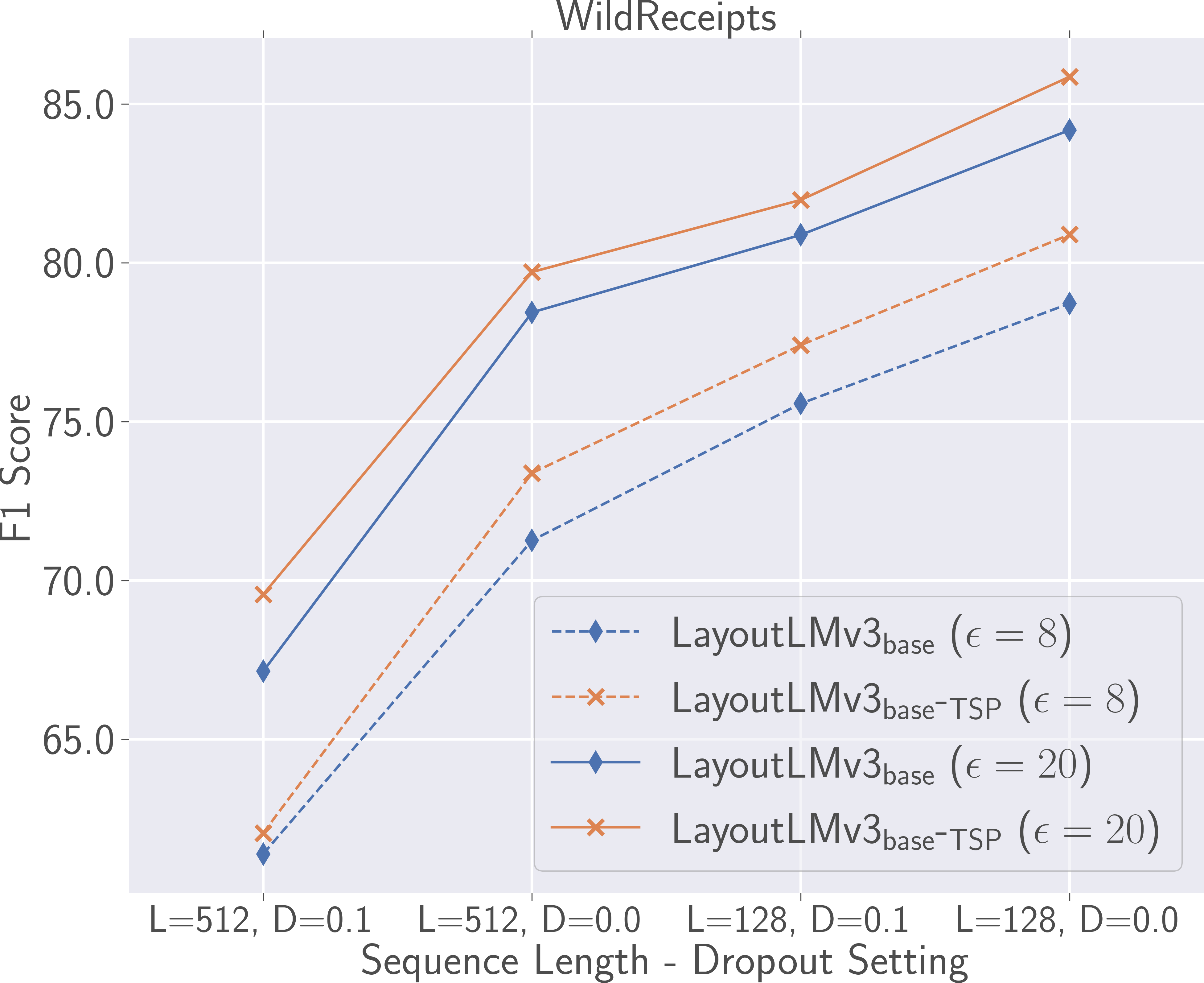}
\end{subfigure}
\begin{subfigure}{0.25\linewidth}
	\includegraphics[width=\linewidth]{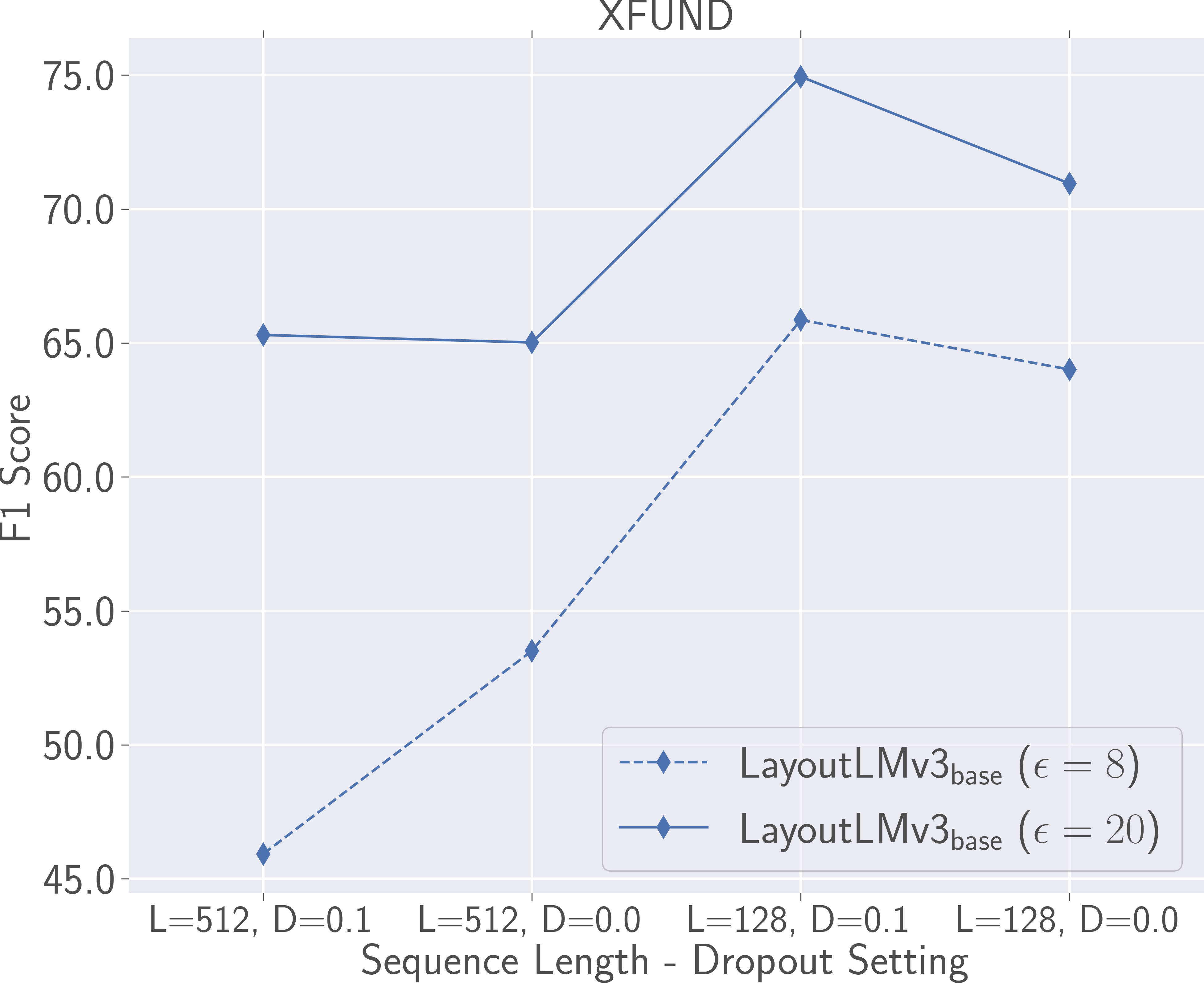}
\end{subfigure}
\caption{DP fine-tuning performance on different datasets under different settings of dropout and sequence length.}
\label{fig:add_dropout_vs_seq_len}
\end{figure*}

\begin{figure}[t!]
	\centering	
	\includegraphics[width=0.8\linewidth]{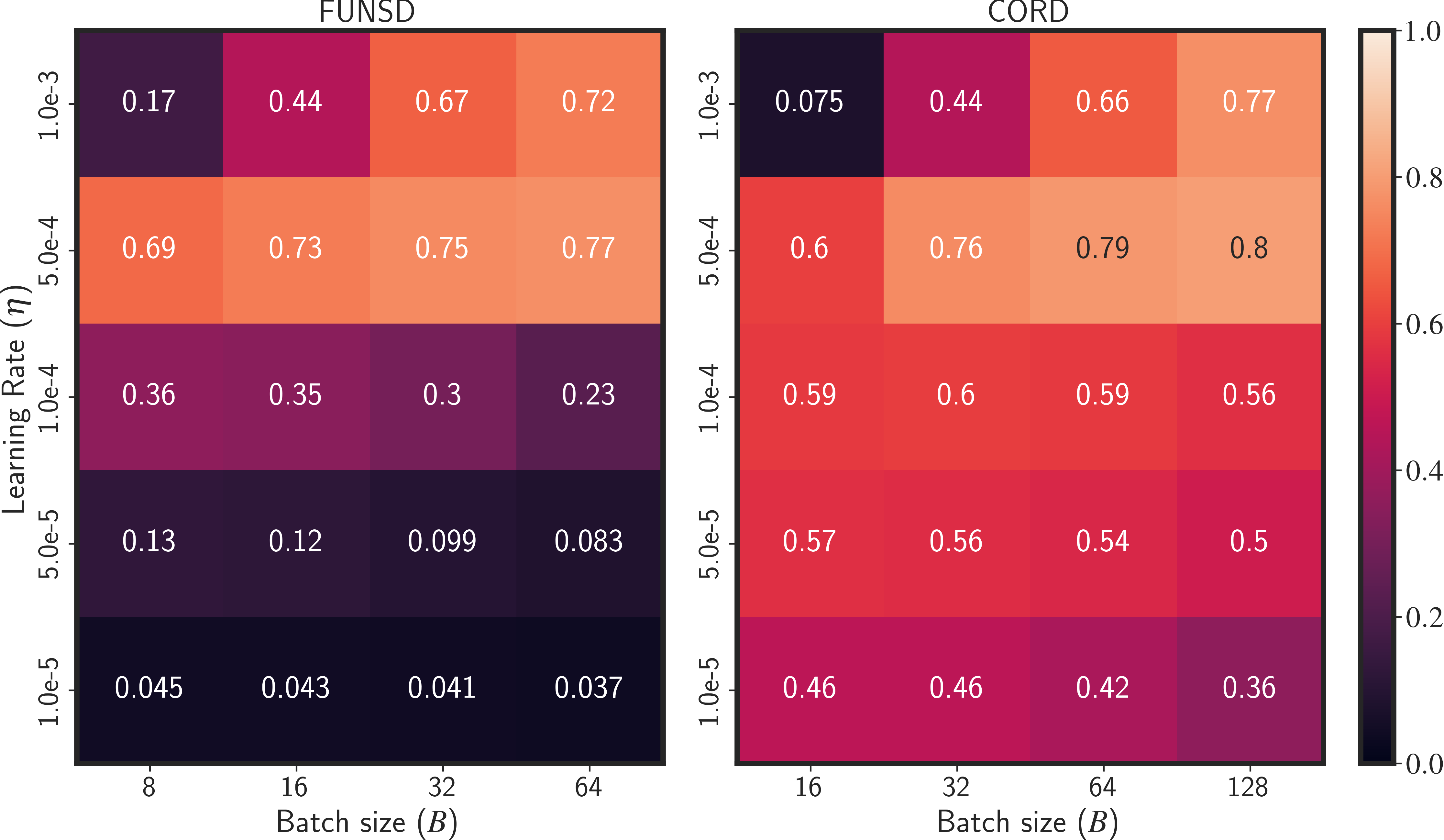}
	\caption{Additional results showing the effect of learning rate on FUNSD and CORD with sequence length $L=512$.}
	\label{fig:add_res_lr_vs_bs}
	\vspace{-1em}
\end{figure}

\begin{figure*}
	\centering	
	\begin{subfigure}{0.2\linewidth}
		\includegraphics[width=\linewidth]{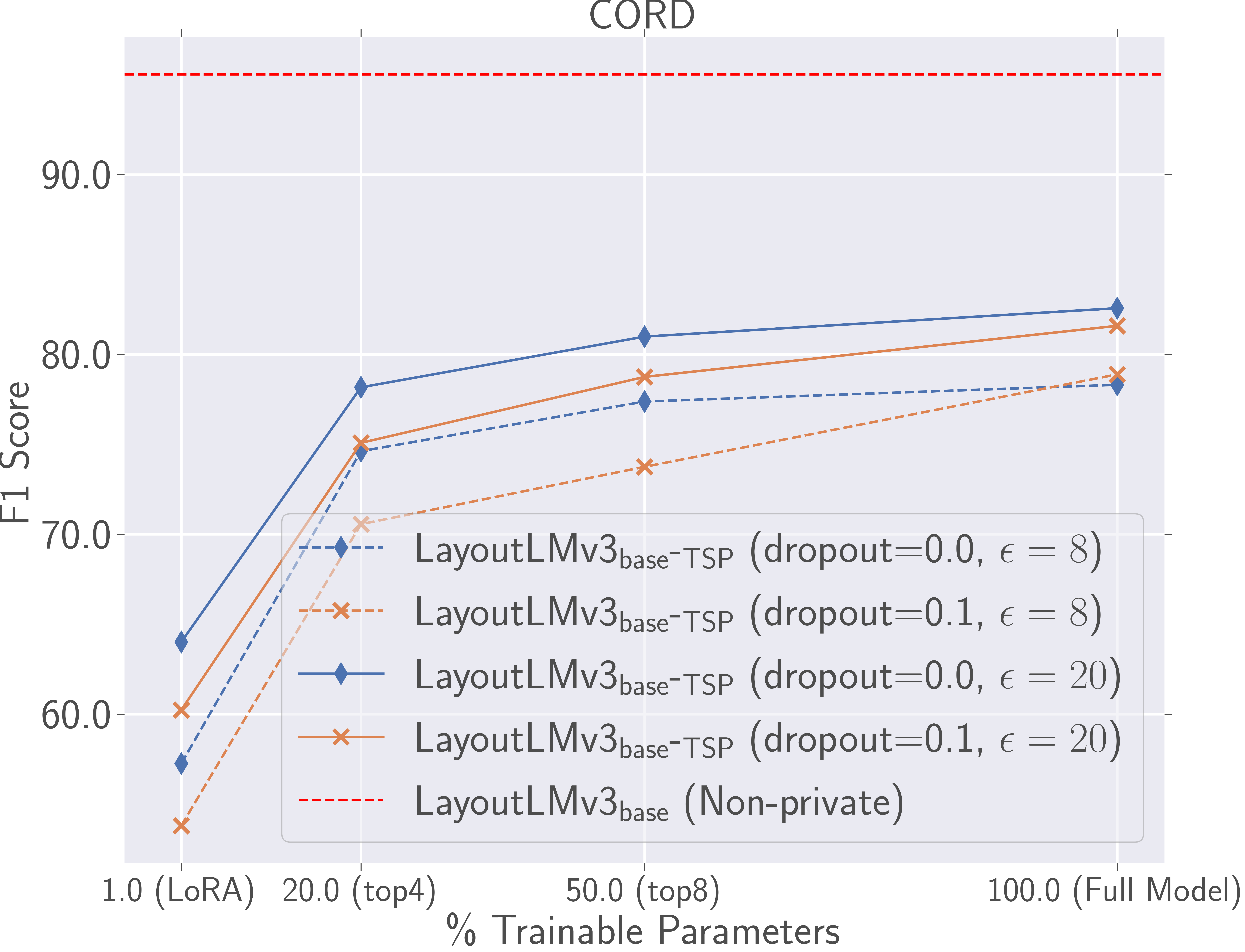}
	\end{subfigure}
	\begin{subfigure}{0.2\linewidth}
		\includegraphics[width=\linewidth]{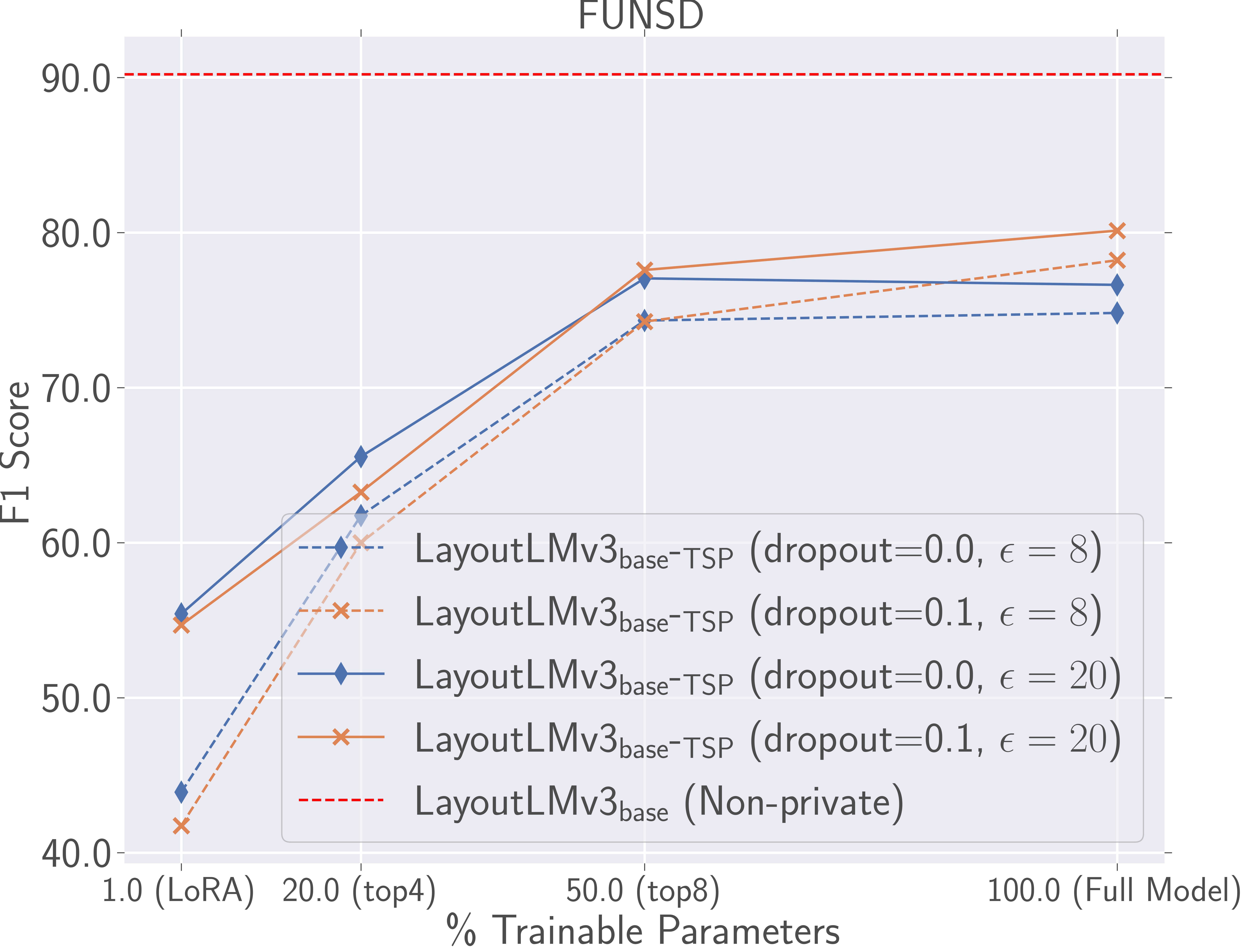}
	\end{subfigure}
	\begin{subfigure}{0.2\linewidth}
		\includegraphics[width=\linewidth]{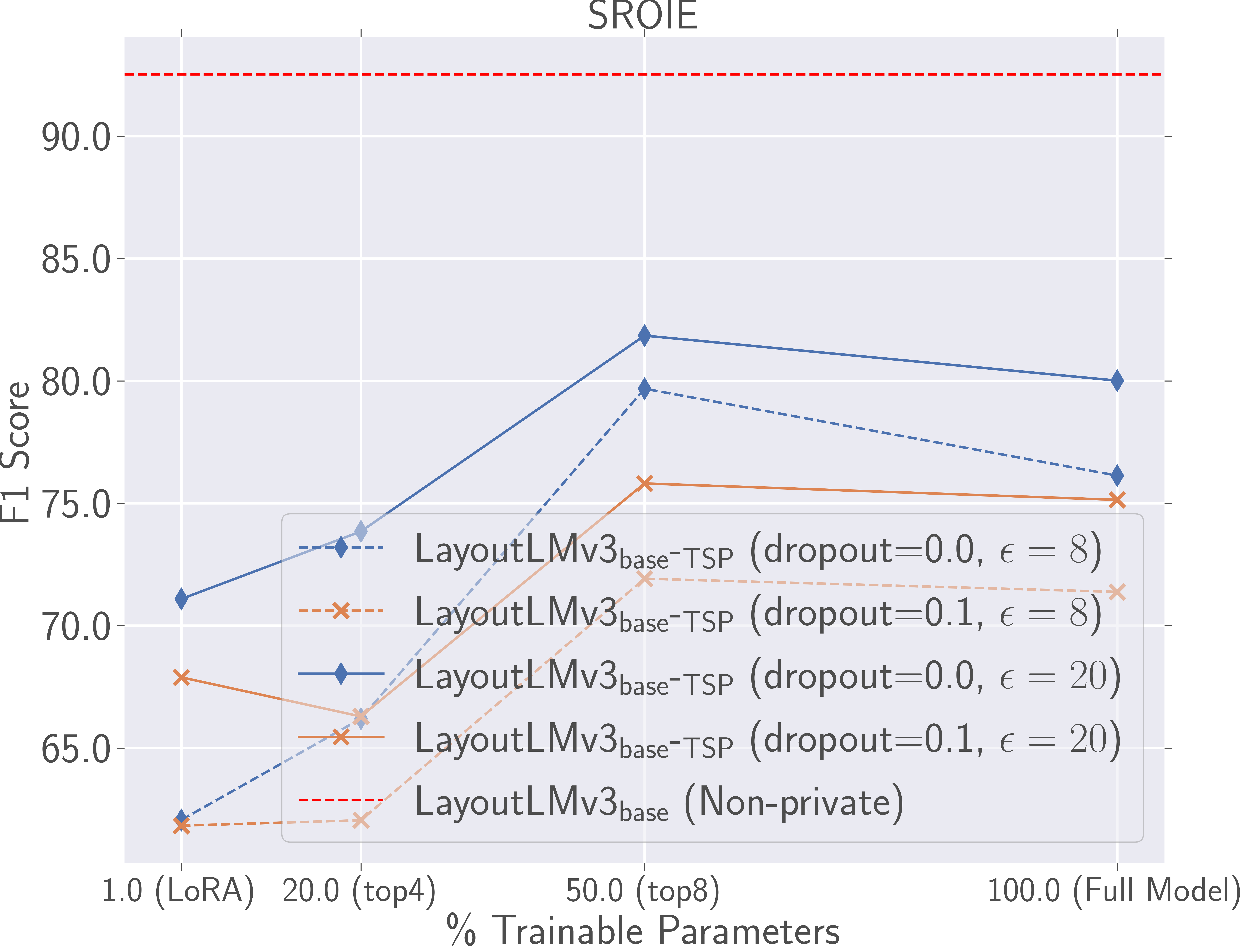}
	\end{subfigure}\\
	\begin{subfigure}{0.2\linewidth}
		\includegraphics[width=\linewidth]{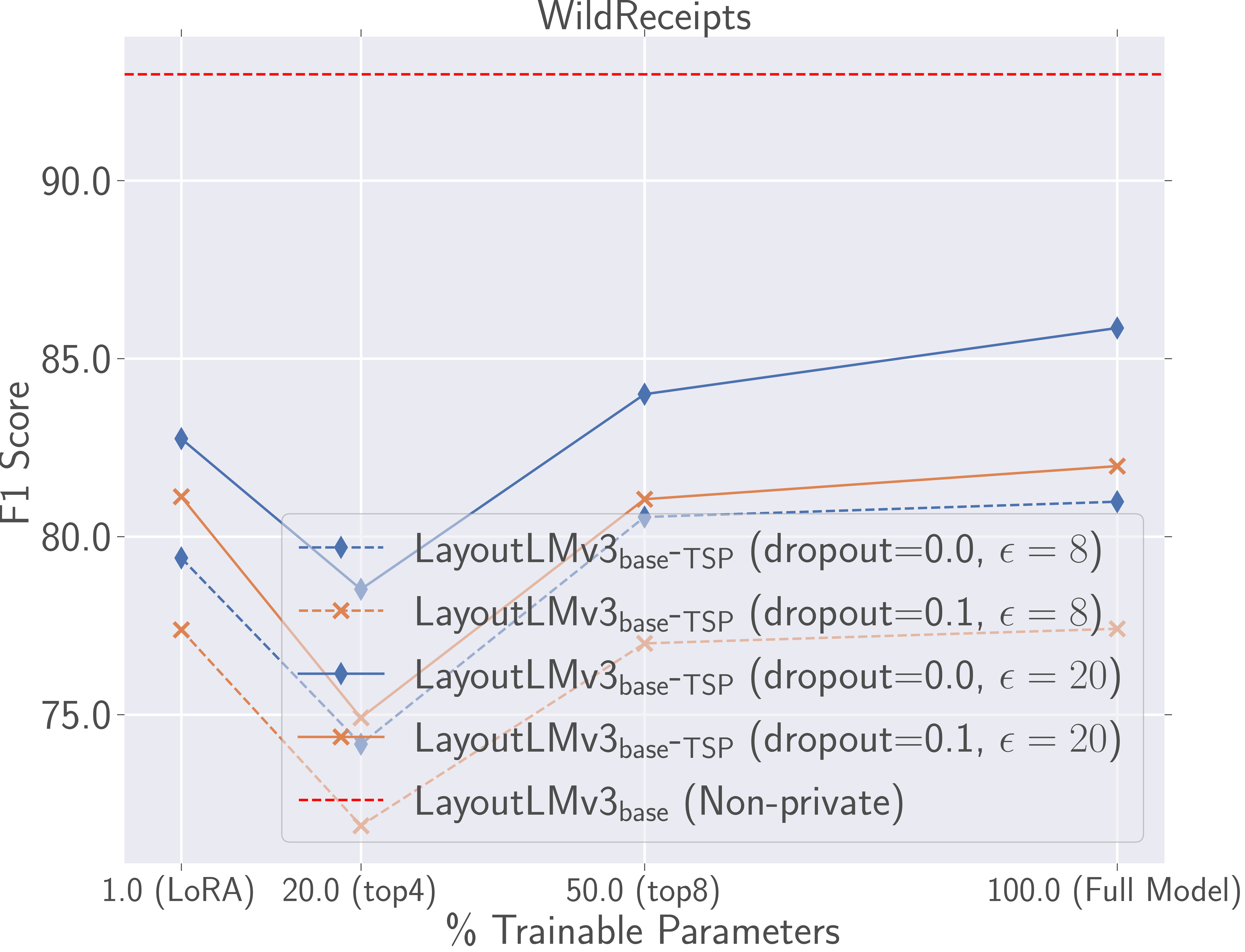}
	\end{subfigure}
	\begin{subfigure}{0.2\linewidth}
		\includegraphics[width=\linewidth]{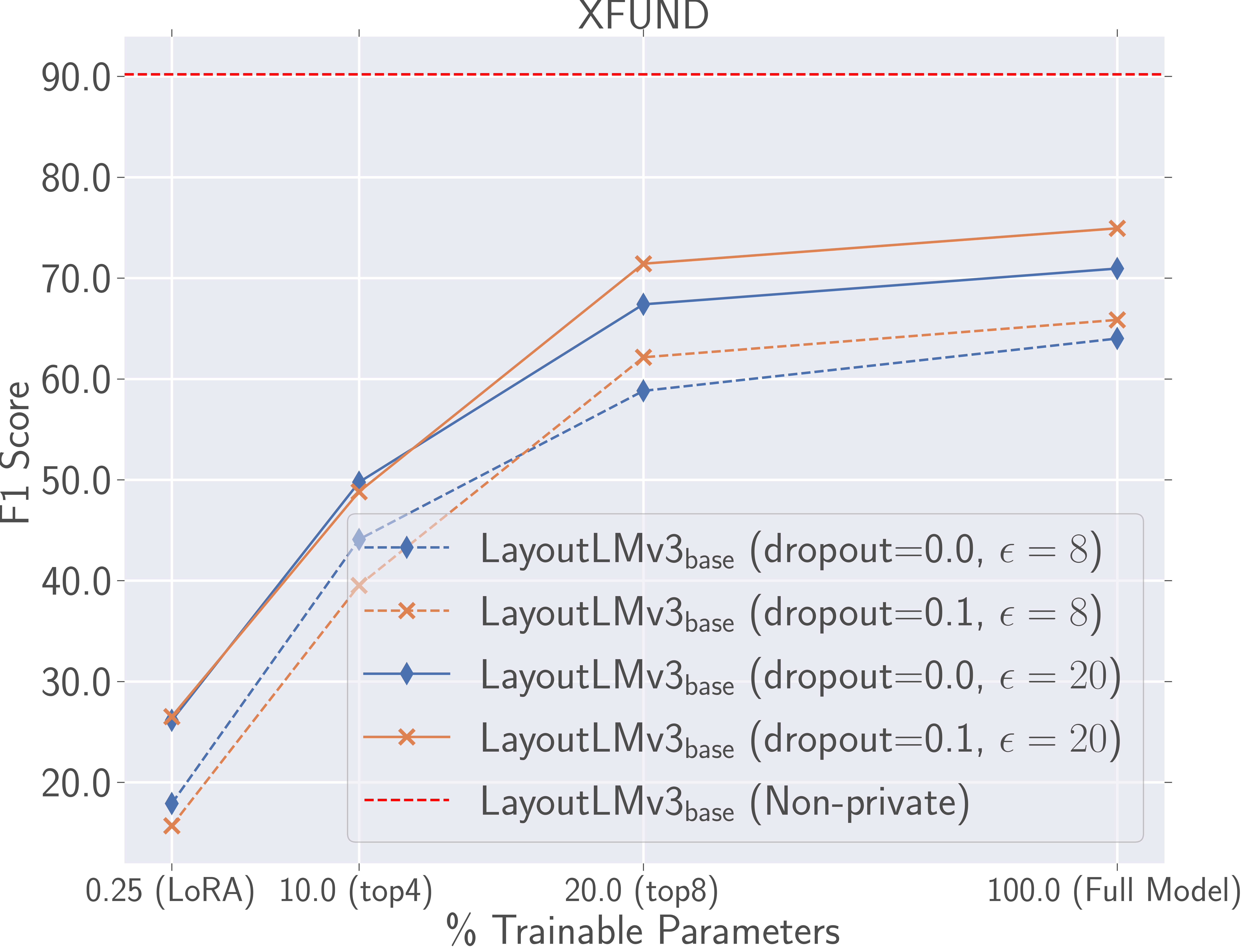}
	\end{subfigure}
	\begin{subfigure}{0.2\linewidth}
		\includegraphics[width=\linewidth]{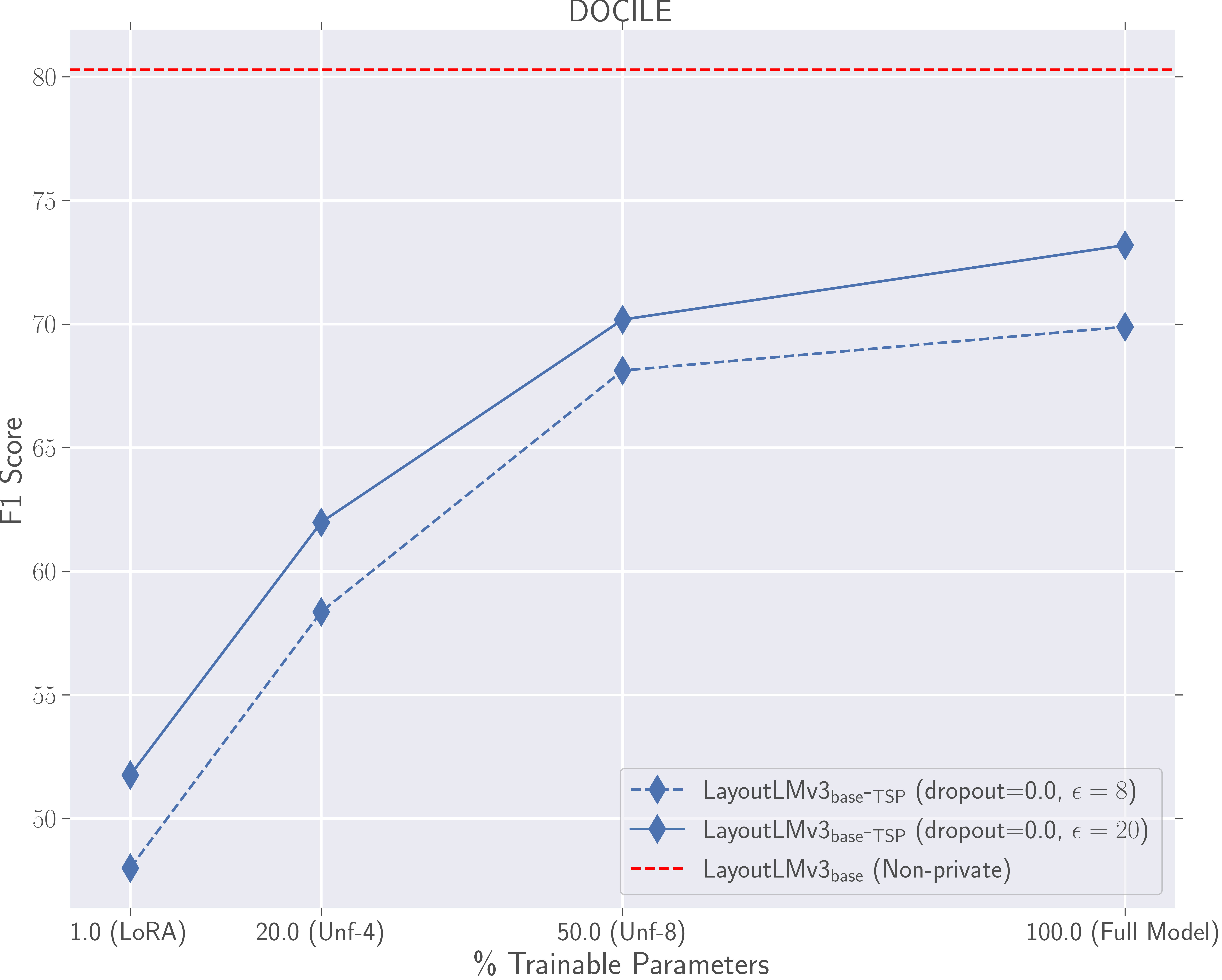}
	\end{subfigure}
	\caption{DP fine-tuning performance on different datasets under different parameter reduction techniques is shown.}
	\label{fig:add_res_param_red}
\end{figure*}
Similarly, WildReceipts had matching entity types to the four target entity types in the SROIE dataset, namely store name, store address, date, and total value. In addition, the DOCILE dataset contained similar entities with the names vendor name, vendor address, date issue, and multiple entities for amount total.
\label{app:dataset_details}
\section{Default Training Hyperparameters}
A summary of the default hyperparameters used in our study is provided in \cref{tab:default-hyperparams}.
\label{app:default_training_params}
\section{Final training hyperparameter configuration}
The final training hyperparameters for each dataset that were used to compute the results for DP and DP-FL scenarios reported in \cref{table:perf-eval} are listed in \cref{tab:default-hyperparams}.
\label{app:final_training_params}
\section{Additional Results - Batch sizes and Learning Rates}
Additional results on the effects of batch size and learning rates on model performance are shown in \cref{fig:add_res_bs} and \cref{fig:add_res_lr_vs_bs}.
\raggedbottom
\label{app:add_res_batch_size}
\section{Additional Results - Dropout and Sequence Lengths}
\label{app:add_res_dropout_seq_len}
Additional results regarding the impact of sequence length and dropout on model performance are presented in \cref{fig:add_dropout_vs_seq_len}.
\section{Additional Results - Parameter Reduction Techniques}
\cref{fig:add_res_param_red} presents additional results regarding the effects of different parameter reduction techniques on model performance. For different datasets, similar results were obtained, with full fine-tuning performing the best and top8 following closely behind. Surprisingly, LoRA performed exceptionally well only on WildReceipts datasets, sometimes achieving performance close to full fine-tuning even with only 1\% of the parameters trained.
\label{app:add_res_param_red}

\section{DP-Adam vs DP-SGD}
\begin{figure}[t!]
	\centering	
	\begin{subfigure}{\linewidth}
		\includegraphics[width=\linewidth]{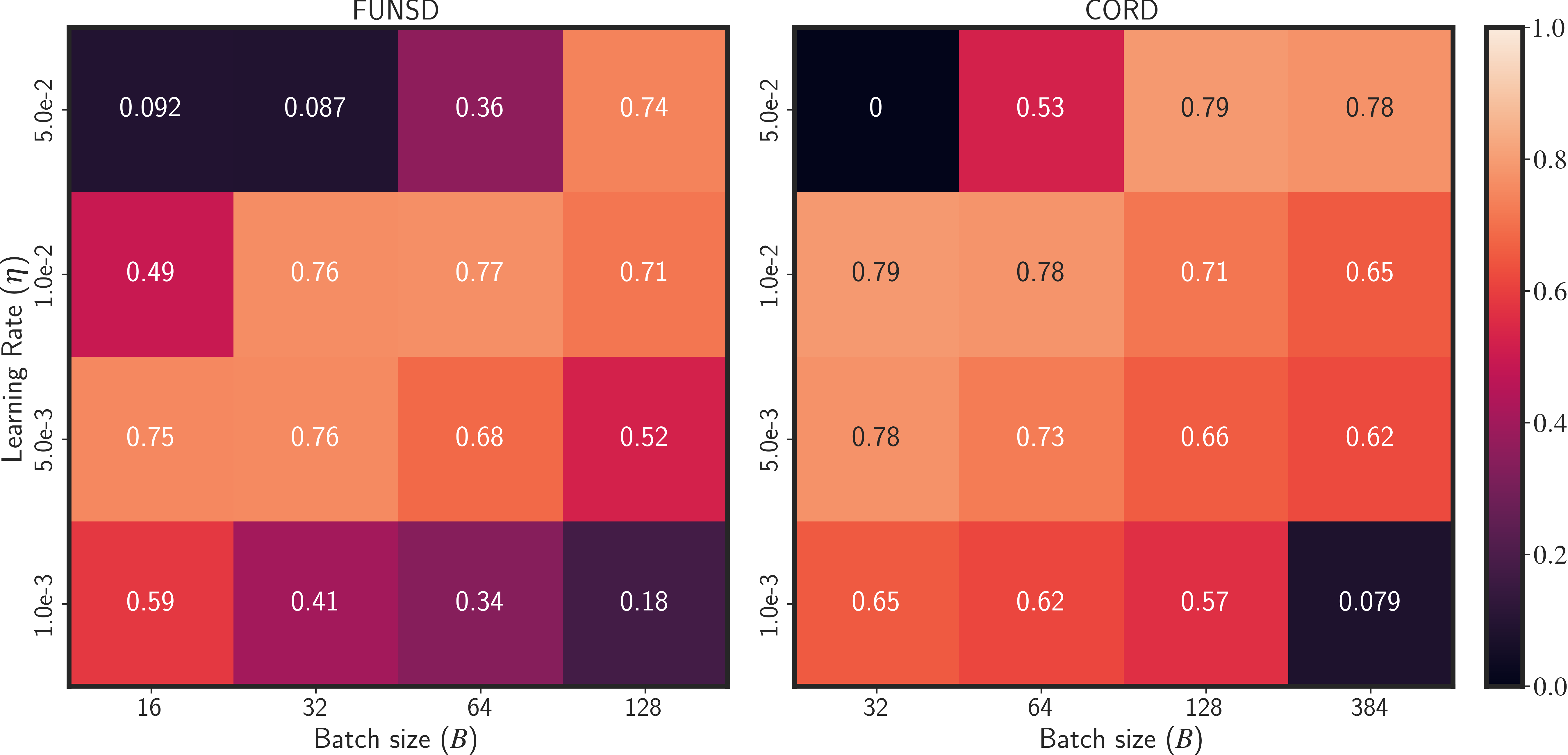}
		\caption{$\epsilon=8$}
	\end{subfigure}\\
	\begin{subfigure}{\linewidth}
		\includegraphics[width=\linewidth]{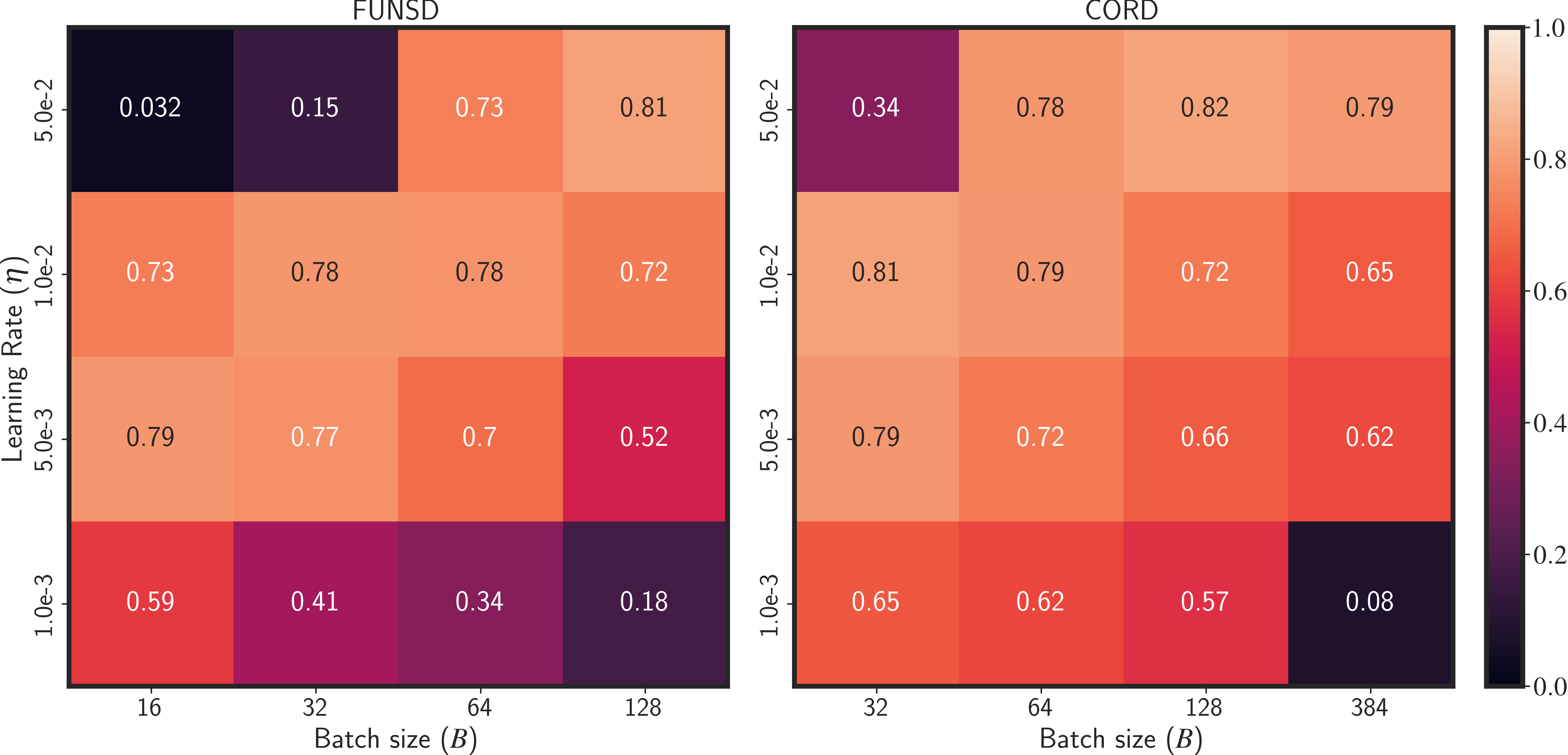}
		\caption{$\epsilon=20$}
	\end{subfigure}
	\caption{DP fine-tuning performance using DP-SGD under varying settings of batch size and learning rates. Results reported here were obtained using the same hyperparameters as used for DP-Adam given in \cref{tab:final-hyperparams} but with SGD optimizer in place of Adam optimizer.}
	\label{fig:dpsgd-vs-dpadam}.
\end{figure}

Our work focused primarily on DP-Adam for differential privacy instead of DP-SGD, mainly because Adam is predominantly used for fine-tuning non-private transformer models. However, it would be worthwhile to investigate the potential of DP-SGD for private fine-tuning as well. In \cref{fig:dpsgd-vs-dpadam}, we present the results of our additional experiments to assess the performance of DP-SGD for private fine-tuning on FUNSD and CORD datasets. As can be observed, for both $\epsilon=8$ and $\epsilon=20$, the performance of DP-SGD was comparable to that of DP-Adam. Furthermore, we observed that DP-SGD was less sensitive to small batch sizes than DP-Adam, and with an appropriate combination of batch size and learning rate, it was able to achieve better performances even with very small batch sizes ($16{\sim}32$).
\end{document}